\pdfoutput=1
\documentclass[10pt,twocolumn,letterpaper]{article}
\usepackage[OT1]{fontenc}
\usepackage[pagenumbers]{cvpr}
\usepackage{cuted}
\usepackage{multirow}
\usepackage{makecell}
\newcommand{\method}{HairDeck\xspace}
\definecolor{cvprblue}{rgb}{0.21,0.49,0.74}
\usepackage[pagebackref,breaklinks,colorlinks,allcolors=cvprblue]{hyperref}

\title{HairDeck: Animation-Ready Stylized Hair Cards from a Single Image}
\author{
Yuze He$^{1,*}$ \quad Yixiang Wei$^{1,*}$ \quad Wang Zhao$^{1}$ \quad Chuhao Chen$^{2}$ \quad Yanning Zhou$^{3,\dagger}$ \quad Yong-Jin Liu$^{1,\dagger}$\\[4pt]
$^{1}$Tsinghua University \quad $^{2}$Carnegie Mellon University \quad $^{3}$The Chinese University of Hong Kong
}

\begin{document}
\maketitle
\begingroup
\renewcommand{\thefootnote}{\fnsymbol{footnote}}
\footnotetext[1]{Equal contribution.}
\footnotetext[2]{Corresponding authors.}
\endgroup
\begin{strip}
\centering
\includegraphics[width=\linewidth]{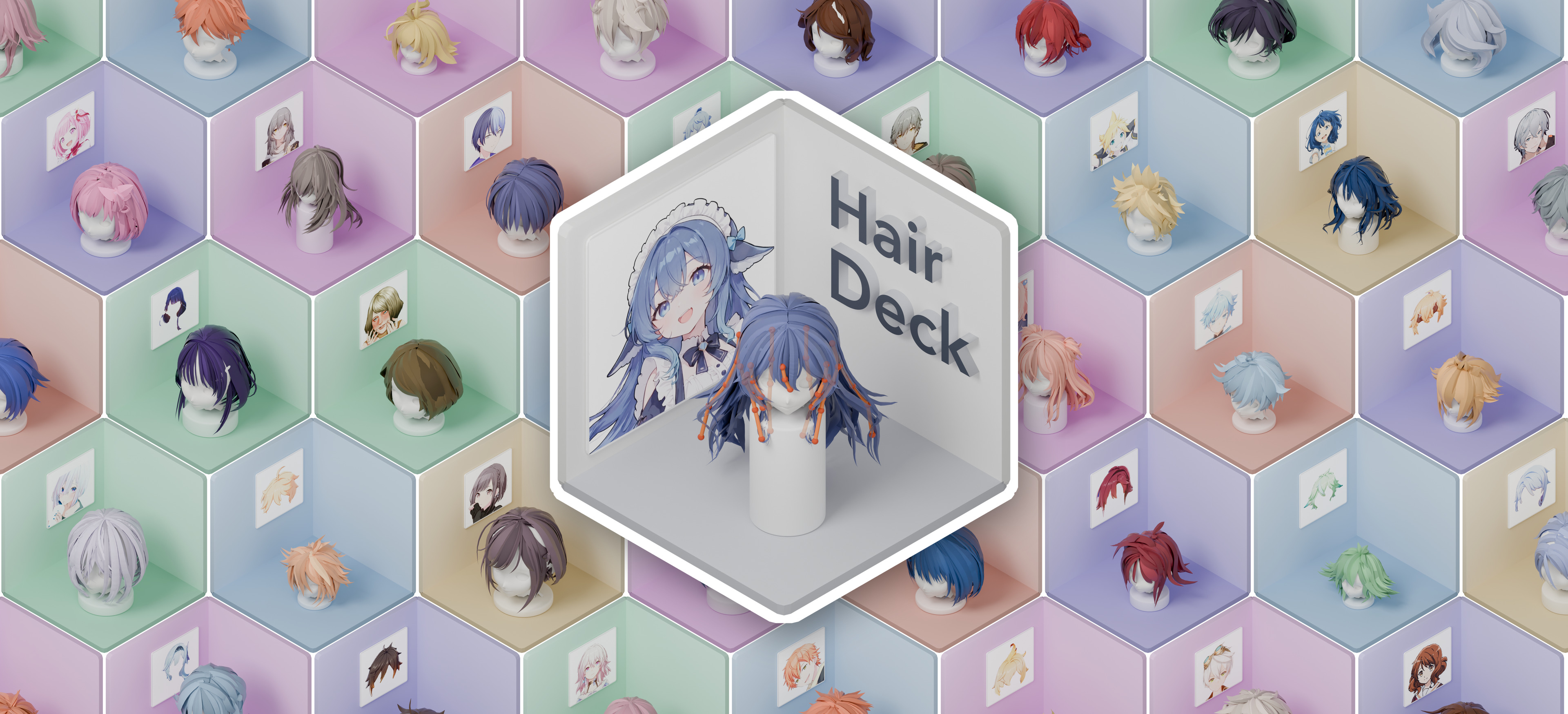}
\captionsetup{hypcap=false}
\captionof{figure}{\method turns a single image into a hairstyle of stylized hair cards, together with the spring-bone chains that animate it. Center: the generated spring-bone chains drawn over the hair, fainter where they run inside it. Hair colors are taken from the input images for display. Videos are on the project page.}
\captionsetup{hypcap=true}
\label{fig:teaser}%
\end{strip}

\begin{abstract}
The hair of stylized characters in games and avatars is built from hair cards: artist-placed ribbons whose geometry is the visible hair, animated by spring bones.
Existing image-based methods produce realistic hair or a single fused mesh, while CHARM, the only generator of stylized hair cards, decodes a hairstyle autoregressively as a long sequence of control points.
Observing that a card hairstyle is an unordered set of artist-defined primitives, we propose \method, which generates it as one: each card is a single token, and a set diffusion transformer denoises all cards together.
From a single image, \method generates the cards and the spring-bone chains that drive them, yielding an animatable asset that exports to ready-to-use downstream formats.
On CHARM's own test set, it lowers Chamfer distance by 44\% and card-orientation error by more than half while running 33 times faster, and from single images it outperforms CHARM, image-to-3D pipelines and strand-based methods.
\par\noindent Project page: \url{https://hyzcluster.github.io/HairDeck}
\end{abstract}

\section{Introduction}
\label{sec:intro}

3D generators now produce convincing shapes from a single image~\cite{zhang2024clay,xiang2024structured,xiang2026trellis2,yang2025hunyuan3d21}, but not the structure that artists build assets from.
The hair of stylized characters in games and avatars is a clear example.
It is made of \emph{hair cards}: a few dozen to over a hundred ribbons, each placed by an artist, whose geometry is the visible hair.
Unlike the alpha-textured hair cards of realistic hair, flat strips whose texture stands in for many strands~\cite{koh2001simple,zheng2025auto,tojo2025strands2cards}, these cards have a rhombic or triangular cross-section whose width and thickness vary along the card.
Such hair is commonly animated with \emph{spring bones}: short chains of bones hung from the head whose tips are simulated every frame, so that the hair sways and settles.
Similar simulations ship as Dynamic Bone in Unity~\cite{dynamicbone}, PhysBones in VRChat~\cite{physbones}, Kawaii Physics in Unreal Engine~\cite{kawaiiphysics} and a node in Godot~\cite{godotspringbone}.
Artists place every card and rig every chain by hand.

Existing methods do not produce this asset.
Hair generators output realistic strands~\cite{sklyarova2023haar,he2024perm,rosu2025difflocks}, including recent ones that accept stylized images~\cite{long2025tangled,luo2026hairgpt,shen2026hairlrm,wang2026hairflow}, and card extraction turns such strands into alpha-textured cards for realistic hair~\cite{zheng2025auto,tojo2025strands2cards}.
Image-to-3D generators~\cite{xiang2026trellis2,yang2025hunyuan3d21} output a single fused mesh with no cards to animate.
CHARM~\cite{he2025charm} is the only generator of stylized hair cards and introduced the benchmark for them.
It generates the control points of a hairstyle one at a time, as a sequence of hundreds to thousands of discretized tokens.

We observe that a card hairstyle is an unordered set of artist-defined primitives, and we generate it as one.
Like CHARM, we describe each card by its control points (position, width and thickness), and we add its cross-section type and its root normal, the width direction at the root, so each card keeps the cross-section and the orientation the artist gave it.
What we change is the unit of generation, with two designs.
First, each card becomes one continuous token: a card VAE encodes its variable-length control points into a latent, and the token holds this latent with the card's root position, root normal and cross-section type, so a hairstyle has one token per card, about 70 on average.
Second, all tokens of a hairstyle are generated together: a set diffusion transformer without positional embedding denoises them with flow matching~\cite{lipman2023flow,liu2023flow}, and an existence flag on each token decides how many cards appear.
Each design makes generation more accurate and faster.
Together they amount to a simple rule for assets made of artist-defined primitives: one token per primitive, and no order among them.

A single image shows the hair relative to the head, so we predict the cards in a head frame anchored to the face, which improves where the hair sits without changing its shape.
Images also come in many drawing styles, so we widen the appearance of our training renders in two ways: synthetic style augmentation, which improves accuracy on stylized inputs, and renders restyled by an image-editing model~\cite{wu2025qwenimage}, which keep the card structure stable across styles.

Cards alone do not move.
We generate the spring-bone chains with the same recipe: each chain becomes one token holding its path, joint count and stiffness, and a second set diffusion transformer generates all chains of a hairstyle together from its card tokens.
With a fixed geometric skinning, the cards and chains form an animatable asset that exports to ready-to-use downstream formats.

In summary, our contributions are:
\begin{itemize}
\item A card token for stylized hair: a representation that keeps the cross-section type and the root normal, and a card VAE that encodes the variable-length control points into one continuous latent and decodes the card by arc length.
\item A set formulation of hair-card generation: a set diffusion transformer without positional embedding generates all card tokens of a hairstyle together, and an existence flag on each token decides how many cards appear.
\item Single-image generation with a face-anchored head frame, synthetic style augmentation and restyled renders.
\item Spring-bone chains generated as a second set conditioned on the card tokens, which makes the output animatable.
\end{itemize}

We evaluate on CHARM's test set with its protocol and also measure the orientation of every card against the original artist meshes.
\method lowers Chamfer distance by 44\% and card-orientation error by more than half while running 33 times faster.
From a single image, it outperforms CHARM, image-to-3D pipelines and strand generators.

\section{Related Work}
\label{sec:related}

\begin{figure*}[t]
\centering
\includegraphics[width=\linewidth]{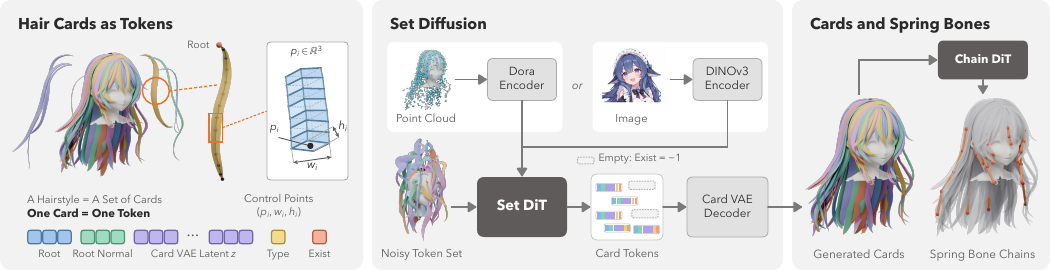}
\caption{Overview. Left: a hairstyle is a set of hair cards. Each card, a polyline of control points with width and thickness, becomes one token holding its root position, root normal, card VAE latent $\mathbf{z}$, cross-section type and existence flag. Middle: the set DiT denoises the token set with flow matching, attending across cards and to the condition, a point cloud or an image. Empty slots let the model decide how many cards to generate. Right: the card VAE decodes the tokens into cards, and the chain DiT generates their spring-bone chains.}
\label{fig:pipeline}
\end{figure*}

\paragraph{Strand-based hair modeling.}
Hair reconstruction methods can recover strands from multi-view images or videos~\cite{nam2019strand,rosu2022neuralstrands,sklyarova2023neural,wu2024monohair,takimoto2024dr,luo2024gaussianhair,zakharov2024human} and from single images~\cite{wu2022neuralhdhair,zheng2023hairstep,zheng2024unified,sklyarova2025im2haircut}.
Generative models synthesize strands from latent codes~\cite{zhou2023groomgen,he2024perm}, text~\cite{sklyarova2023haar} and images~\cite{rosu2025difflocks}.
Recent ones accept stylized or anime images~\cite{long2025tangled,luo2026hairgpt,shen2026hairlrm,wang2026hairflow}, but their output is still strands, the representation of realistic hair.
The stylized hair we generate uses cards instead.

\paragraph{Hair cards.}
Real-time engines render realistic hair with textured strips~\cite{koh2001simple}, the alpha-textured hair cards whose texture stands in for many strands.
Zheng~\etal~\cite{zheng2025auto} and Strands2Cards~\cite{tojo2025strands2cards} convert strand grooms into such cards, HairLRM~\cite{shen2026hairlrm} applies the former to its generated strands, CGHair~\cite{luo2026cghair} reconstructs card-based hair from multi-view images, Lee~\etal~\cite{lee2025generating} inpaint strips from the sparse strands of a single portrait with a diffusion model on the scalp UV map, HairCS~\cite{lu2026haircs} turns cards back into strands, and production tools ship card generators for grooms~\cite{uehaircard,houdinihaircard}.
Stylized hair uses a different card, whose geometry is the visible hair.
CHARM~\cite{he2025charm} introduced the generation of such cards with a control-point representation, a dataset of anime hairstyles (the benchmark we use) and an autoregressive transformer that emits the control points of a hairstyle one at a time.
We add the cross-section type and the root normal to its description of a card and change the unit of generation.

\paragraph{Stylized 3D characters.}
Image-based methods for anime characters reconstruct stylized heads~\cite{chen2023panic} or full characters~\cite{peng2024charactergen,wang2024nova}, and others separate the body, clothing and hair into components~\cite{he2024stdgen,he2026stdgenpp,qian2026animeready}.
In all of them the hair is a continuous surface, fused with the character or as a separate component, rather than a set of cards.
Earlier work modeled anime and cartoon hair from sketches, skeletons or parametric clumps~\cite{mao2004sketch,sakai2013skeleton,yamamoto2017npr} or as 2.5D layers~\cite{yeh20142}.

\paragraph{Structured and set generation.}
Native 3D generators encode shapes as latent sets~\cite{zhang20233dshape2vecset,zhao2023michelangelo,zhang2024clay,chen2025dora} or structured latents~\cite{xiang2024structured,xiang2026trellis2} and output a single surface.
Assets built from discrete elements are mostly generated autoregressively: meshes as sequences of face or vertex tokens~\cite{siddiqui2024meshgpt,chen2024meshanythingv2,weng2024scaling,zhao2025deepmesh,hao2024meshtron}, ordered along triangle strips and UV charts in Strips as Tokens~\cite{xu2026sato}, and shapes as sequences of primitives~\cite{ye2025primitiveanything}.
Other work generates the elements of an asset together.
3DTopia-XL~\cite{chen20253dtopiaxl} gives each primitive one token and denoises the tokens with a diffusion transformer without positional encoding.
StrokeFusion~\cite{zhou2026strokefusion} encodes each sketch stroke with a VAE and diffuses the unordered set of stroke latents with presence flags.
DiffuScene~\cite{tang2024diffuscene}, SALAD~\cite{koo2023salad}, MeshFlow~\cite{sun2026meshflow}, BrepGen~\cite{xu2024brepgen} and PartCrafter~\cite{lin2025partcrafter} jointly generate the objects of a scene, part latents, the triangles of a mesh, the faces of CAD models and the parts of objects.
We build on this set formulation and design it for hair cards: a card VAE turns each card into one token, the set is conditioned on a point cloud or a single image, and a second set of spring-bone chains is conditioned on the cards.
On the same tokens, set generation is both more accurate and faster than autoregressive generation.

\paragraph{Rigging and secondary motion.}
Automatic rigging predicts a skeleton and skinning weights for a given mesh~\cite{xu2020rignet,song2025magicarticulate,liu2025riganything,guo2025makeitanimatable,song2025puppeteer,sun2026animator}.
UniRig~\cite{zhang2025unirig} also predicts spring bones and their physical attributes on VRoid characters~\cite{vroidHub}, and DRiVE~\cite{sun2025drive} generates skeletons that include hair and skirt bones for anime characters represented as 3D Gaussians.
Both rig an existing asset.
We generate the spring-bone chains together with the hair they drive, conditioned on the generated cards, and simulate them with the spring-bone update of the VRM specification~\cite{vrmspec0,univrm}.

\section{Method}
\label{sec:method}

\begin{figure*}[t]
\centering
\includegraphics[width=\linewidth]{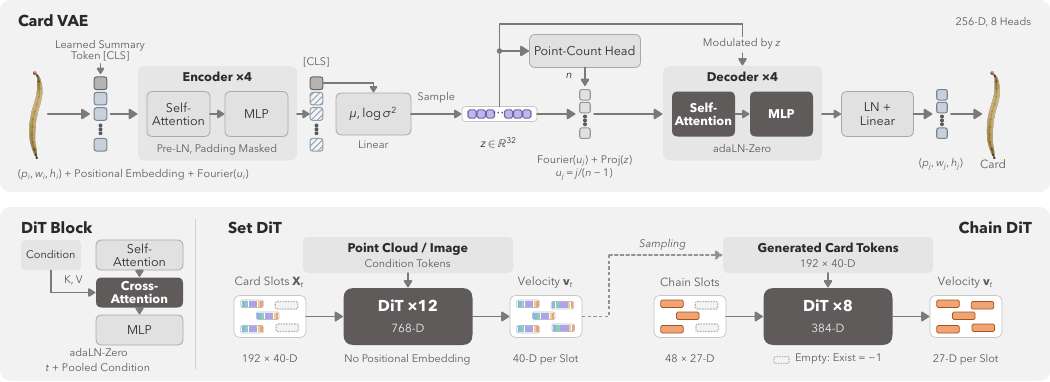}
\caption{Network design. Top: the card VAE. The encoder reads the control points of one card with their arc-length fractions $u_i$, plus a learned summary token \texttt{[CLS]}, whose output alone gives the mean and log-variance of the 32-d latent $\mathbf{z}$. The decoder predicts the number of points $n$ and evaluates the card at $n$ uniformly spaced arc-length fractions, modulated by $\mathbf{z}$ (adaLN-Zero). Bottom: both DiTs use the same block design (left), modulated by the timestep and the pooled condition. The set DiT predicts the velocity of 192 card token slots without positional embedding, conditioned on a point cloud or an image. The chain DiT does the same for 48 spring-bone chain slots, conditioned on the card tokens sampled by the set DiT (dashed arrow). Empty slots (dashed boxes) have existence $-1$, filled ones +1.}
\label{fig:arch}
\end{figure*}

We propose \method, which generates the cards of a hairstyle and the spring-bone chains to animate them (\cref{fig:pipeline}).
Each card is encoded into one continuous token (\cref{sec:card_token}), and a set diffusion transformer (set DiT) generates all card tokens together, conditioned on a single image (\cref{sec:image}) or on a point cloud (\cref{sec:set_dit}), the input of CHARM's benchmark.
The card VAE decodes each token back into control points, from which we build the mesh.
A second set DiT, the chain DiT, generates the spring-bone chains conditioned on the card tokens, and a fixed geometric skinning attaches the cards to them (\cref{sec:chains}).

\subsection{Hair cards as tokens}
\label{sec:card_token}

\paragraph{Cards.}
Like CHARM~\cite{he2025charm}, we describe a hair card by a polyline of $n$ control points $(\mathbf{p}_i, w_i, h_i)$, each with a position $\mathbf{p}_i \in \mathbb{R}^3$, a width $w_i$ and a thickness $h_i$.
We add two attributes that this description leaves out: the cross-section type $s$, rhombic or triangular, instead of one rhombic cross-section for all cards, and the root normal $\mathbf{d}_1$, the width direction of the card at its root, read from the artist mesh instead of being estimated from the first three control points.
To build a card mesh, we propagate $\mathbf{d}_1$ along the polyline with a rotation-minimizing (Bishop) frame~\cite{bishop1975there}, giving the width direction $\mathbf{d}_i$ at every control point, and sweep the cross-section along it.

\paragraph{Card VAE.}
Cards differ in their number of control points, while the generator needs a token of fixed size.
A card VAE encodes the control points of a card, relative to its root position $\mathbf{p}_1$, into a latent $\mathbf{z} \in \mathbb{R}^{32}$ (\cref{fig:arch}, top).
Its encoder is a transformer over the control points, each tagged with its arc-length fraction $u_i \in [0, 1]$, together with a learned summary token whose output gives the mean and log-variance of $\mathbf{z}$.
The decoder predicts the number of control points $n$ from $\mathbf{z}$ with a classification head and evaluates the card at $n$ uniformly spaced arc-length fractions, embedded with low-frequency Fourier features~\cite{tancik2020fourier}, with a transformer modulated by $\mathbf{z}$ through adaLN-Zero~\cite{Peebles2022DiT}.
The loss is a squared error on positions, widths and thicknesses in world units, plus a cross-entropy on $n$ and a small KL term~\cite{kingma2014auto}.

\paragraph{Card tokens.}
Each card becomes a token
\begin{equation}
\mathbf{x} = (\mathbf{p}_1, \mathbf{d}_1, \mathbf{z}, s, e) \in \mathbb{R}^{40}
\label{eq:token}
\end{equation}
that holds its root position and root normal, its latent, its cross-section type $s$ ($-1$ for rhombic, $+1$ for triangular) and an existence flag $e = +1$.
A hairstyle is a set of $N = 192$ token slots, and an empty slot is all zeros except $e = -1$.
After sampling, we read $s$ and $e$ by their sign.

\subsection{Set diffusion of hair cards}
\label{sec:set_dit}

\paragraph{Set diffusion transformer.}
The set DiT, a diffusion transformer~\cite{Peebles2022DiT}, denoises the $N$ token slots of a hairstyle together (\cref{fig:arch}, bottom).
It uses no positional embedding, so it is equivariant to permutations of the slots, and no order of the cards is ever defined.
Each block applies self-attention across the slots, cross-attention to the condition tokens $\mathbf{C}$ and an MLP, all modulated by the timestep and the pooled condition through adaLN-Zero.
For a point cloud, $\mathbf{C}$ is the output of a frozen Dora-VAE encoder~\cite{chen2025dora}; for an image, it comes from an image encoder (\cref{sec:image}).

\paragraph{Flow matching.}
We train the generator with flow matching~\cite{lipman2023flow,liu2023flow}.
For the token set $\mathbf{X}_0$ of a training hairstyle and Gaussian noise $\boldsymbol{\epsilon}$, the network $\mathbf{v}_\theta$ predicts the velocity along the straight path between them:
\begin{align}
\mathbf{X}_t &= (1 - t)\,\mathbf{X}_0 + t\,\boldsymbol{\epsilon}, \label{eq:path}\\
\mathcal{L} &= \mathbb{E}_{\mathbf{X}_0, \boldsymbol{\epsilon}, t}\,\big\| \mathbf{v}_\theta(\mathbf{X}_t, t, \mathbf{C}) - (\boldsymbol{\epsilon} - \mathbf{X}_0) \big\|_{\mathbf{w}}^2, \label{eq:fm}
\end{align}
where $\|{\cdot}\|_{\mathbf{w}}$ weights the root position and the existence flag by 4 and the other components by 1.
We drop the condition in training to allow classifier-free guidance~\cite{ho2022classifier}.
At test time we integrate with the Euler method and guidance and decode every slot with $e > 0$.
The existence flags let the card count settle during denoising together with the shapes.

\begin{figure*}[t]
\centering
\includegraphics[width=\linewidth]{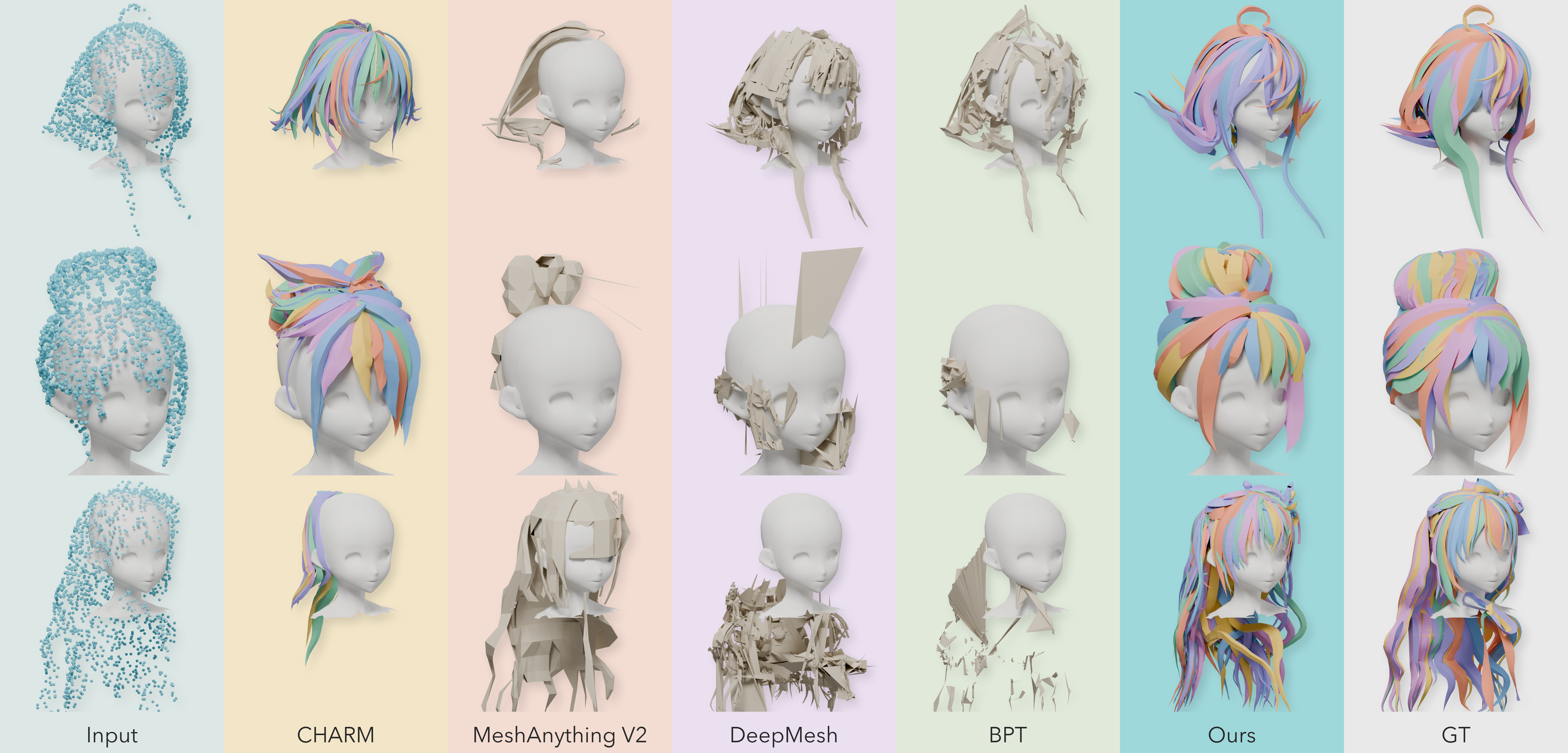}
\caption{Point clouds from CHARM's test set. CHARM and \method output hair cards, colored per card like the artist ground truth; MeshAnything V2, DeepMesh and BPT output meshes. All results are shown on a fixed head at each character's placement.}
\label{fig:cmp_pc}
\end{figure*}

\subsection{Generation from a single image}
\label{sec:image}

\paragraph{Image condition.}
The image model is a set DiT of the same architecture, conditioned on the output tokens of a frozen DINOv3 ViT-L/16~\cite{simeoni2026dinov3} applied to a square crop around the hair and the face.
For in-the-wild images, SAM~3~\cite{carion2026sam3} provides the hair and face masks.

\begin{figure}[t]
\centering
\includegraphics[width=0.9\linewidth]{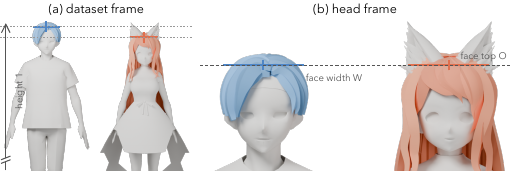}
\caption{Head frame. (a) The dataset frame scales every character to height 1, so heads differ in size and height; the right character's ears count toward its height and lower its head. (b) In the head frame, all characters share the face width $W$ and face top $O$.}
\label{fig:supp_headframe}
\end{figure}

\paragraph{Head frame.}
The size and position of a hairstyle follow its head, whereas the dataset frame scales every character to unit height.
We therefore predict the cards in a head frame anchored to the face (\cref{fig:supp_headframe}).
Each character is mapped by $\mathbf{p} \mapsto k\,\mathbf{p} + \mathbf{b}$, where $k = W / w$ scales its face width $w$ to the median face width $W$ of the training set and $\mathbf{b}$ moves the top center of its face to a common point $O$; widths and thicknesses scale with $k$.
On rendered test characters, we map the output back with each character's own $k$ and $\mathbf{b}$, which places the hair on its real head.

\paragraph{Training images.}
We render every training character from 8 views in Blender~\cite{blender}, with random body poses, hair motion from spring-bone simulation, and cameras from the front to the back of the head; the target is always the rest-pose card set.
We widen the appearance of the renders with synthetic style augmentation, such as halftone, XDoG~\cite{winnemoller2012xdog} line drawings, flat colors and painterly filters (\cref{sec:supp_impl}), and we mix in renders restyled into three drawing styles, gacha (mobile-game) art, thick paint and semi-realistic painting, by an image-editing model~\cite{wu2025qwenimage}.

\begin{figure*}[t]
\centering
\includegraphics[width=\linewidth]{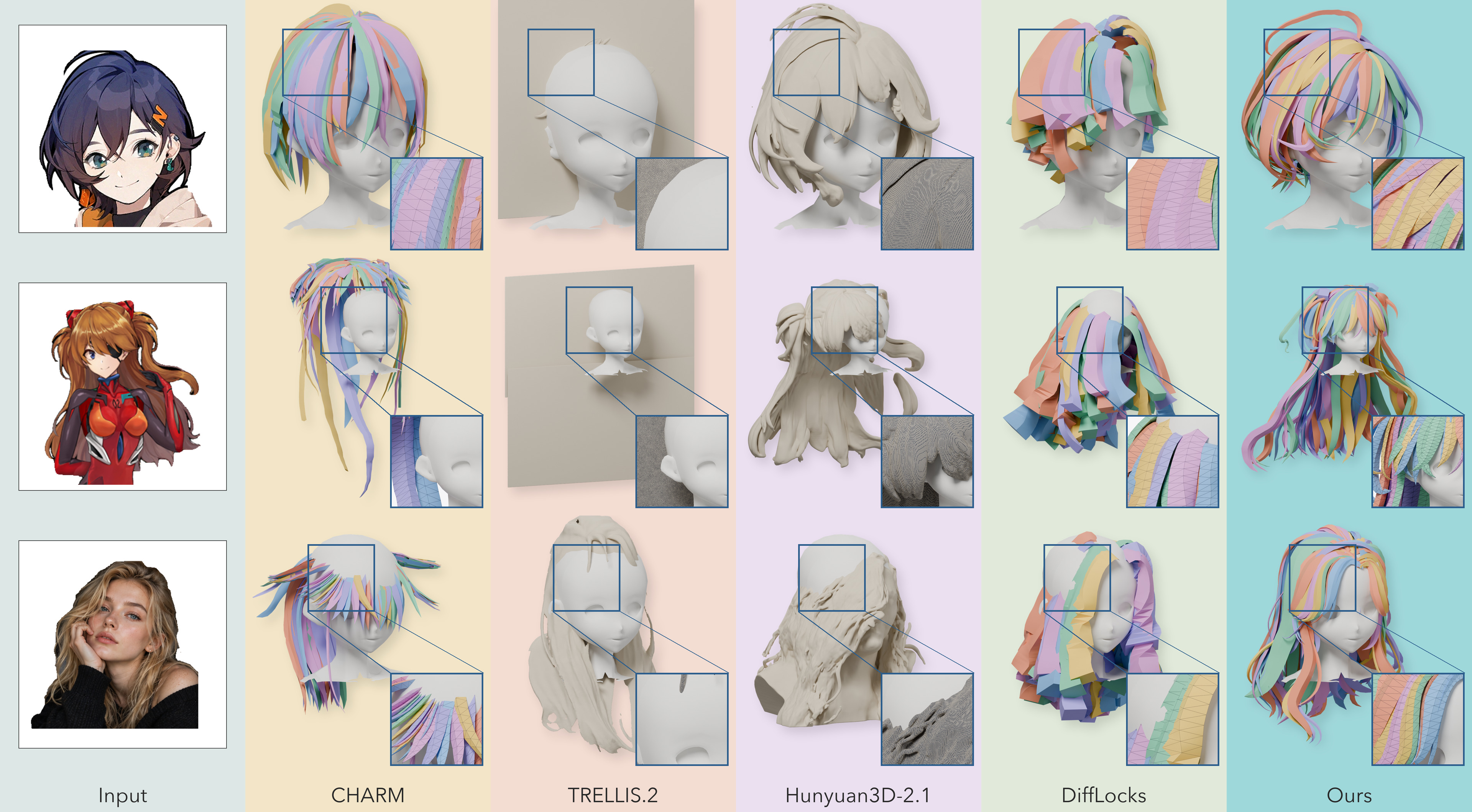}
\caption{Single images: two in-the-wild illustrations (top) and an AI-generated photorealistic portrait (bottom). Insets show the outlined regions with mesh edges. TRELLIS.2 and Hunyuan3D-2.1 take the hair region as input, and TRELLIS.2 outputs a flat plate facing the camera for both illustrations. Each card has its own color.}
\label{fig:cmp_img}
\end{figure*}

\subsection{Spring-bone chains}
\label{sec:chains}

Cards alone are static, but the hair of game and avatar characters must move with them.
Artists achieve this with spring-bone chains, sequences of joints hung from the head whose simulated motion the cards follow, and rig them by hand for every hairstyle; we generate the chains so that the output can be animated directly.

\paragraph{Chain tokens.}
A chain becomes a 27-dimensional token of its path (8 points by arc length), joint count, stiffness and existence flag, and a hairstyle a set of up to 48 chains.

\paragraph{Chain generation.}
The chain DiT has the same block design as the set DiT and generates the chain tokens of a hairstyle in the head frame with flow matching, cross-attending to its card tokens (\cref{fig:arch}).
It is trained on the ground-truth card tokens and, at test time, reads the generated ones; cards from a point cloud are first mapped into the head frame (\cref{sec:supp_impl}).

\paragraph{Skinning and simulation.}
After generation, a fixed geometric skinning binds the cards to the chains.
We place the joints uniformly along each generated path.
A card follows the chain closest to the lower 60\% of its centerline if their mean distance is below 0.4 face widths, and otherwise moves rigidly with the head; each of its points follows the nearest segment of that chain, blended linearly with the neighboring segment within half a segment of each joint, with a weight that rises from 0 at the chain's root to 1 over 0.05 face widths of arc length.
We tune these two thresholds against the artists' skin weights in the training data.
We simulate the chains with the VRM spring-bone update of UniVRM~\cite{vrmspec0,univrm}, using the predicted stiffness and the drag and gravity of most VRoid chains.

\subsection{Implementation details}
\label{sec:impl}
The card VAE has four encoder and four decoder layers of width 256, the set DiT 12 layers of width 768 (181.3M parameters) and the chain DiT 8 layers of width 384 (30.5M).
We train the point-cloud model for 80K steps with batch size 64 and the image model for 100K steps with batch size 48, both with AdamW~\cite{loshchilov2019decoupled} at learning rate $10^{-4}$; the image model starts from the point-cloud model.
We sample with 50 Euler steps and a guidance scale of 2; \cref{sec:supp_impl} lists the remaining settings.

\section{Experiments}
\label{sec:experiments}

\begin{figure*}[t]
\centering
\includegraphics[width=\linewidth]{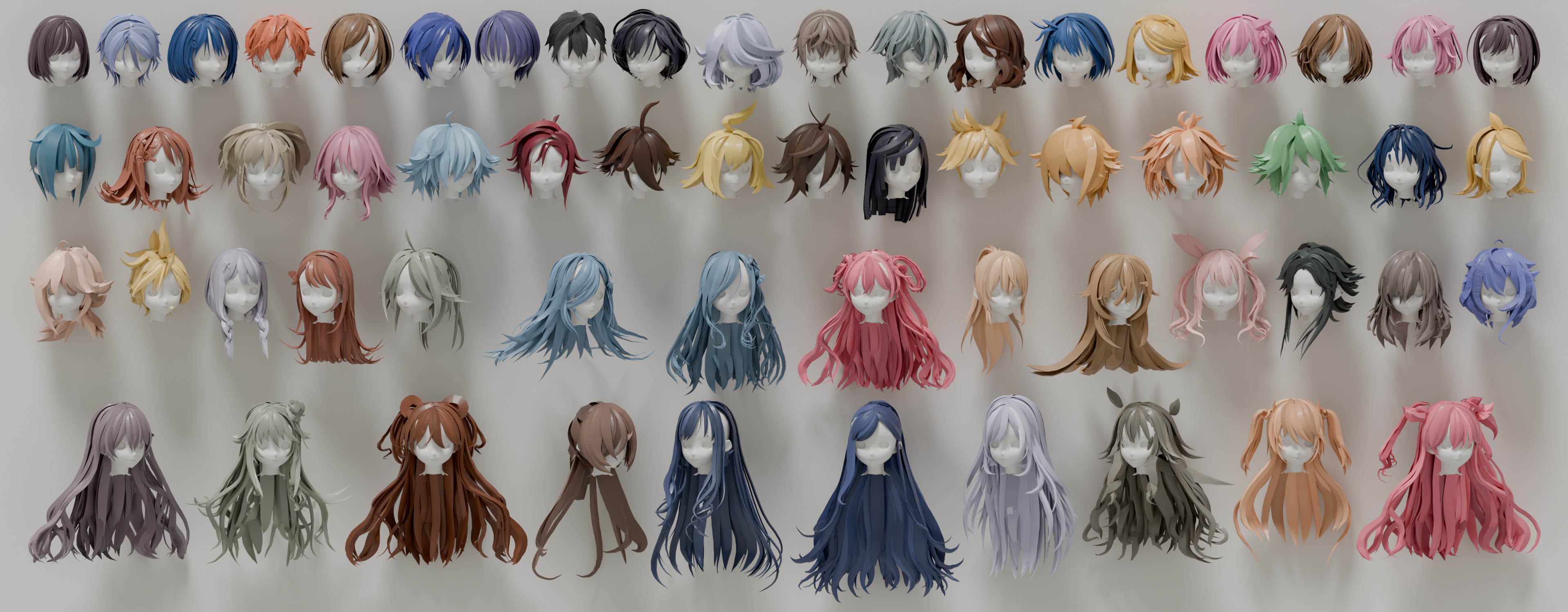}
\caption{Hairstyles generated by \method from in-the-wild images. Hair colors are taken from the images for display.}
\label{fig:wall}
\end{figure*}

\subsection{Setup}
\label{sec:setup}

\paragraph{Data.}
We use the dataset of CHARM~\cite{he2025charm}, built from VRoid characters~\cite{vroidHub}: we take 36,440 of its training hairstyles and test on its 100 test hairstyles, with the point clouds CHARM released.
For single images, we render the front views of the 76 test characters whose complete models are available, and we collect 40 in-the-wild images, anime illustrations sampled at random and stratified by hairstyle.

\paragraph{Metrics.}
On point clouds we follow CHARM's protocol: it compares the generated cards with its ground truth, rebuilt from its card templates, by Chamfer distance (CD), earth mover's distance (EMD), Hausdorff distance (HD) and voxel IoU.
As the templates estimate card orientations, we also evaluate against the original artist meshes.
Our orientation metric matches every generated card to the artist card closest in CD and averages the angle between their width directions over 32 positions along the card.
Root precision and recall are the shares of generated and ground-truth cards whose root position lies within 2\% of the character height of a root position in the other set.
For single images, we align every output to the ground truth (\cref{sec:supp_baselines}) and report the aligned CD and IoU on surface samples.
The failure rate is the share of results that have fewer than 25\% of the ground-truth cards (16 cards for in-the-wild images) or leave the scalp bare (hair covers less than 80\% of the top or the back of the head); for the image-to-3D generators, whose output is one fused surface without cards, only the scalp criterion applies.
We time every method on one NVIDIA RTX A6000 GPU (48\,GB) with batch size 1, from the input to the cards, before meshing.

\subsection{Generation from point clouds}
\label{sec:exp_pc}

\begin{table}[t]
\centering
\caption{Generation from point clouds on CHARM's 100 test hairstyles, with its protocol; orientation on the original meshes (artist cards meshed from root normals: 7.5\textdegree); time per hairstyle. Means leave out 1 DeepMesh and 3 BPT outputs that fail to scan.}
\label{tab:pc}
\footnotesize
\setlength{\tabcolsep}{2.7pt}
\begin{tabular}{@{}lcccccc@{}}
\toprule
Method & CD$\downarrow$ & EMD$\downarrow$ & HD$\downarrow$ & IoU$\uparrow$ & Orient.$\downarrow$ & Time$\downarrow$ \\
\midrule
MeshAnything V2~\cite{chen2024meshanythingv2} & 0.0166 & 0.0182 & 0.0630 & 0.671 & -- & 176\,s \\
DeepMesh~\cite{zhao2025deepmesh} & 0.0357 & 0.0381 & 0.1001 & 0.530 & -- & 1050\,s \\
BPT~\cite{weng2024scaling} & 0.0446 & 0.0507 & 0.1063 & 0.451 & -- & 351\,s \\
CHARM~\cite{he2025charm} & 0.0117 & 0.0128 & 0.0498 & 0.753 & 40.0\textdegree & 33.7\,s \\
Retrieval & 0.0119 & 0.0119 & 0.0505 & 0.732 & 24.8\textdegree & -- \\
\midrule
Card-level AR & \underline{0.0078} & \underline{0.0093} & \underline{0.0356} & \underline{0.823} & \textbf{16.6}\textdegree & \underline{6.91}\,s \\
\method & \textbf{0.0066} & \textbf{0.0084} & \textbf{0.0242} & \textbf{0.867} & \underline{16.8}\textdegree & \textbf{1.01}\,s \\
\bottomrule
\end{tabular}
\end{table}

\Cref{tab:pc,fig:cmp_pc} compare \method with CHARM and autoregressive mesh generators on CHARM's test set.
\method lowers CHARM's CD from 0.0117 to 0.0066 and the orientation error from 40.0\textdegree{} to 16.8\textdegree{}, while taking 1.01\,s instead of 33.7\,s.
Retrieving the training hairstyle with the closest condition reaches a CD of 0.0119, so CHARM is about as accurate as recalling a training hairstyle.
The mesh generators output a few large shells (MeshAnything V2, BPT) or triangle fragments (DeepMesh).

The card-level autoregressive model in \cref{tab:pc} generates our card tokens one after another, with about as many parameters as our set DiT.
One token per card already lowers CHARM's CD to 0.0078 and the orientation error to 16.6\textdegree.
Generating the same tokens as a set lowers CD further to 0.0066 and runs 6.8 times faster than the autoregressive model, which also stops early, with 60.8 cards on average against 68.9 for the set DiT and 67.4 in the ground truth.

\subsection{Generation from single images}
\label{sec:exp_img}

\begin{figure*}[t]
\centering
\includegraphics[width=\linewidth]{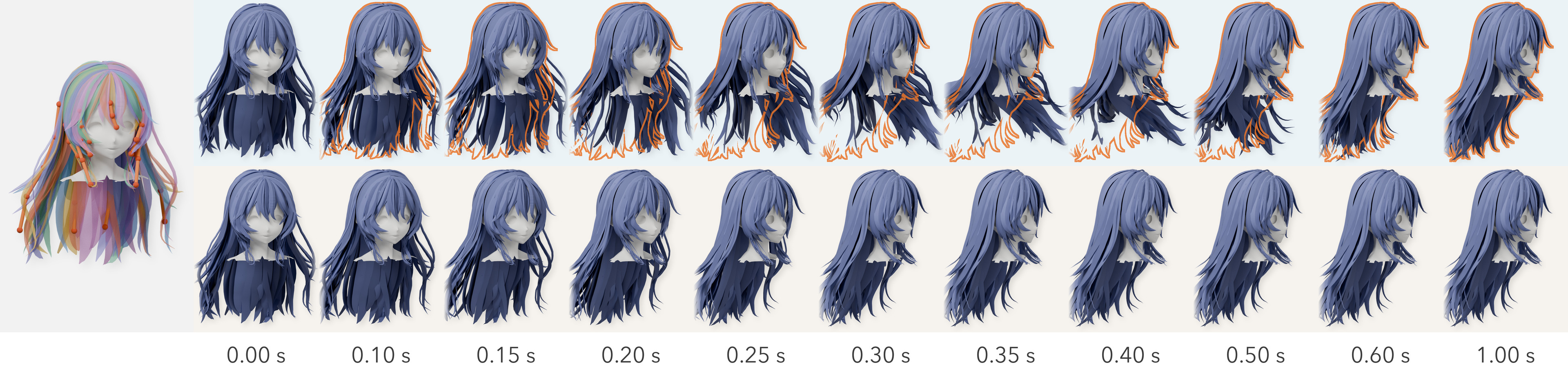}
\caption{Generated spring bones in motion. Left: the generated cards (transparent) with their generated spring-bone chains. Right: a 50\textdegree{} head turn; top row with the chains, bottom row without them. The orange outline marks the hair without spring bones in the same frame.}
\label{fig:spring}
\end{figure*}

\begin{table}[t]
\centering
\caption{Generation from single images: aligned CD and IoU on 76 test characters, failure rates on them and on 40 in-the-wild images (\cref{sec:setup}), and time per image.}
\label{tab:img}
\footnotesize
\setlength{\tabcolsep}{2.6pt}
\begin{tabular}{@{}lccccc@{}}
\toprule
& \multicolumn{2}{c}{Test characters} & \multicolumn{2}{c}{Failure rate$\downarrow$} & \\
\cmidrule(lr){2-3}\cmidrule(lr){4-5}
Method & CD$\downarrow$ & IoU$\uparrow$ & Test & In the wild & Time$\downarrow$ \\
\midrule
CHARM~\cite{he2025charm} & 0.0186 & 0.540 & 34.2\% & 50.0\% & 67.7\,s \\
TRELLIS.2~\cite{xiang2026trellis2} & \underline{0.0139} & \underline{0.638} & \underline{13.2\%} & 77.5\% & 83.0\,s \\
Hunyuan3D-2.1~\cite{yang2025hunyuan3d21} & 0.0148 & \underline{0.638} & 30.3\% & 75.0\% & 60.8\,s \\
DiffLocks~\cite{rosu2025difflocks} + cards & 0.0174 & 0.592 & 17.1\% & \underline{17.5\%} & \underline{15.6}\,s \\
\midrule
\method & \textbf{0.0089} & \textbf{0.725} & \textbf{3.5\%} & \textbf{10.0\%} & \textbf{1.33}\,s \\
\bottomrule
\end{tabular}
\end{table}

\Cref{tab:img} compares \method with CHARM's image model, the image-to-3D generators TRELLIS.2~\cite{xiang2026trellis2} and Hunyuan3D-2.1~\cite{yang2025hunyuan3d21}, and DiffLocks~\cite{rosu2025difflocks}, whose strands we convert into cards.
On the rendered test characters, \method has the lowest aligned CD (0.0089, against 0.0139 to 0.0186); it fails on 3.5\% of them and on 10\% of the in-the-wild images, against 13.2\% to 34.2\% and 17.5\% to 77.5\% for the other methods, and takes 1.33\,s per image, against 15.6\,s to 83.0\,s.
Cascades that pass a point cloud sampled from an image-to-3D output to CHARM or our point-cloud model reach aligned CDs of 0.0138 to 0.0170 and fail on 22.4\% to 32.9\% of the test characters (\cref{tab:supp_cascade}).
\Cref{fig:cmp_img} covers anime illustrations and a photorealistic portrait: the image-to-3D generators produce one fused surface, and the converted DiffLocks strands leave the crown bare and end in blocky clumps on the portrait.
\Cref{fig:wall} shows hairstyles generated from in-the-wild images in many drawing styles.

\subsection{Ablations}
\label{sec:ablations}

\begin{table}[t]
\centering
\caption{Ablations on point clouds; each row changes one part of \method (all rows in \cref{tab:supp_abl_pc}). Raw control points: the card resampled to 64 points by arc length. Turn: largest angle between consecutive segments of a card, median per hairstyle (artist cards: 14.1\textdegree).}
\label{tab:abl_pc}
\footnotesize
\setlength{\tabcolsep}{3.5pt}
\begin{tabular}{@{}lcccc@{}}
\toprule
Variant & CD$\downarrow$ & IoU$\uparrow$ & Orient.$\downarrow$ & Turn \\
\midrule
\method & \textbf{0.0066} & \textbf{0.867} & 16.8\textdegree & 14.2\textdegree \\
\midrule
raw control points as tokens & 0.0068 & 0.856 & 17.1\textdegree & 21.9\textdegree \\
VAE with index decoding & 0.0079 & 0.823 & 18.1\textdegree & 25.3\textdegree \\
w/o root normal & 0.0070 & 0.850 & 26.4\textdegree & 13.9\textdegree \\
w/o existence (card-count head) & 0.0068 & 0.859 & \textbf{16.7}\textdegree & 14.5\textdegree \\
CHARM's card templates & 0.0068 & 0.856 & 26.7\textdegree & 15.8\textdegree \\
CHARM's encoder (Michelangelo) & 0.0068 & 0.853 & 17.2\textdegree & 14.0\textdegree \\
small model (54.6M) & 0.0072 & 0.847 & 17.5\textdegree & 14.2\textdegree \\
\bottomrule
\end{tabular}
\end{table}

\paragraph{Card token.}
The card VAE latent gives a lower CD than raw control points (\cref{tab:abl_pc}; 0.0066 against 0.0068), and its output is as smooth as the artists' cards (turn 14.2\textdegree{} against 14.1\textdegree), whereas raw control points give jagged cards (21.9\textdegree).
A VAE that decodes the control points by index, as StrokeFusion~\cite{zhou2026strokefusion} decodes the points of a stroke, reaches a CD of 0.0079 with jagged cards (25.3\textdegree); our arc-length decoding, adaLN modulation and world-unit loss make the difference (\cref{tab:supp_vae}).
Without the root normal, the orientation error rises from 16.8\textdegree{} to 26.4\textdegree.
Without the existence flag, a head that predicts the card count first generates too few cards (62.8 against 68.9).
\paragraph{Data, encoder and size.}
Training on CHARM's card templates raises the orientation error to 26.7\textdegree.
With CHARM's encoder (Michelangelo~\cite{zhao2023michelangelo}) or a generator of 54.6M parameters, fewer than CHARM's 61.2M, CD stays at 0.0068 and 0.0072, far below CHARM's 0.0117.

\paragraph{Image model.}
We test the image model on the front views of the 76 test characters and on their restyles into gacha art, thick paint and semi-realistic painting (\cref{fig:supp_style}).
Without the head frame, the model must infer the head's size and position: its CD before alignment rises from 0.0136 to 0.0189 and the overlap (IoU) of the reprojected hair with the input's hair drops from 0.829 to 0.798, while its aligned CD stays at 0.0089.
On the restyles, synthetic style augmentation lowers the aligned CD from 0.0100 to 0.0097, and restyled training images raise root precision from 0.781 to 0.849 at a similar root recall: without them, the model adds cards that start where no artist card does.

\begin{figure*}[t]
\centering
\includegraphics[width=\linewidth]{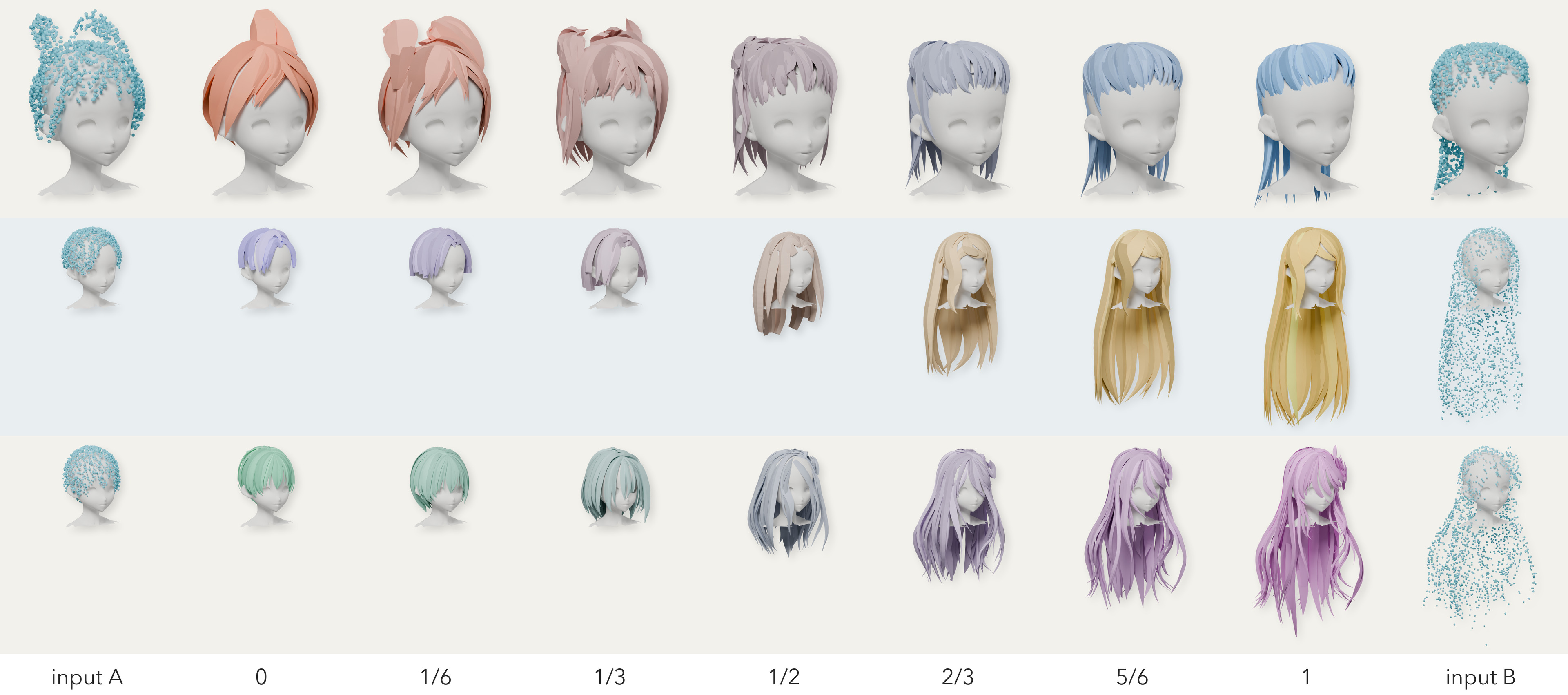}
\caption{Interpolation between two test point clouds, inputs A and B: the condition tokens of B are matched to those of A and blended with the weight $\alpha$ under each column, and all samples share the initial noise. Hair colors and head placements are blended for display.}
\label{fig:supp_interp}
\end{figure*}

\subsection{Spring bones}
\label{sec:exp_spring}

\begin{table}[t]
\centering
\caption{Spring-bone chains on the 58 test hairstyles with chains, against the artists' chains (metrics in \cref{sec:exp_spring}; Ampl.: mean per-point motion, 0.0157\,$W$ for the artists' chains). Bottom rows: generated cards; --: not measured.}
\label{tab:spring}
\footnotesize
\setlength{\tabcolsep}{2.6pt}
\begin{tabular}{@{}lcccccc@{}}
\toprule
Chains & CD$\downarrow$ & Rec.$\uparrow$ & Prec.$\uparrow$ & Motion$\uparrow$ & Ampl. & Fit$\uparrow$ \\
\midrule
Procedural & 0.504 & 0.397 & 0.110 & 0.538 & 0.0473 & 0.832 \\
Retrieval & 0.431 & 0.289 & 0.285 & 0.622 & 0.0157 & 0.811 \\
Ours, artist cards & \textbf{0.279} & \textbf{0.625} & \textbf{0.585} & \textbf{0.702} & 0.0167 & \textbf{0.953} \\
\midrule
Ours, point cloud & 0.319 & 0.542 & 0.543 & -- & -- & 0.952 \\
Ours, single image & 0.407 & 0.390 & 0.402 & -- & -- & 0.949 \\
\bottomrule
\end{tabular}
\end{table}

We evaluate chains on the 58 test hairstyles that have them (\cref{tab:spring,fig:spring}) against the artists' chains: by position (CD in face widths $W$, recall and precision at 0.1\,$W$), by the hair motion under a head turn, nod, jump and sway, and by the fit to the cards (\cref{sec:supp_impl}).
On the artists' cards, the chain DiT recovers the artists' chains better than retrieval (recall 0.625 against 0.289, precision 0.585 against 0.285) and moves the hair more like the artists' chains do: the correlation of the per-point motion is 0.702, against 0.622 for retrieval and 0.538 for a procedural rig that places chains by rules (\cref{sec:supp_baselines}).
On generated cards, recall is 0.542 from point clouds and 0.390 from single images, and the fit to the cards stays at 0.95.

\subsection{Applications}
\label{sec:exp_apps}

\paragraph{Interpolation.}
The Dora encoder computes the condition tokens of a point cloud at query points that farthest point sampling picks from it, so the tokens of two point clouds have no correspondence.
To interpolate between two conditions, we match the tokens of the second to those of the first by Hungarian matching on the positions of their query points and blend the matched tokens linearly (\cref{fig:supp_interp}).

\paragraph{Rigging existing hairstyles.}
The chain DiT reads only card tokens, so it can also rig card hairstyles that an artist modeled without spring bones; its evaluation on the artists' cards in \cref{sec:exp_spring} measures this use.

\section{Limitations and Conclusion}
\label{sec:conclusion}

\paragraph{Limitations.}
\method learns from stylized VRoid hairstyles, so very short hair and the alpha-textured cards of realistic hair lie outside its training data.
It generates the geometry and UV coordinates of the cards but no textures.
Some VRoid characters model bows and hats with hair cards, so \method generates them as cards when the input shows them; filtering such cards from the training data would avoid this.
The chain DiT follows the chain layouts common among VRoid characters and generates fewer chains for unusual ones, such as many chains radiating from the crown.
\Cref{fig:supp_fail} shows failure cases.

\paragraph{Conclusion.}
\method generates the stylized hair cards of a character and the spring-bone chains that animate them from a single image.
Treating a hairstyle as an unordered set of artist-defined primitives, with one token per card and all tokens generated together, makes generation more accurate and faster than autoregressive decoding, and the same recipe generates the chains.
Assets built from other artist-defined primitives, such as garments sewn from panels or drawings made of strokes, can be generated in the same way.

{
    \small
    \bibliographystyle{ieeenat_fullname}
    \bibliography{main}
}

\clearpage
\appendix
\begin{strip}
\centering
\includegraphics[width=\linewidth]{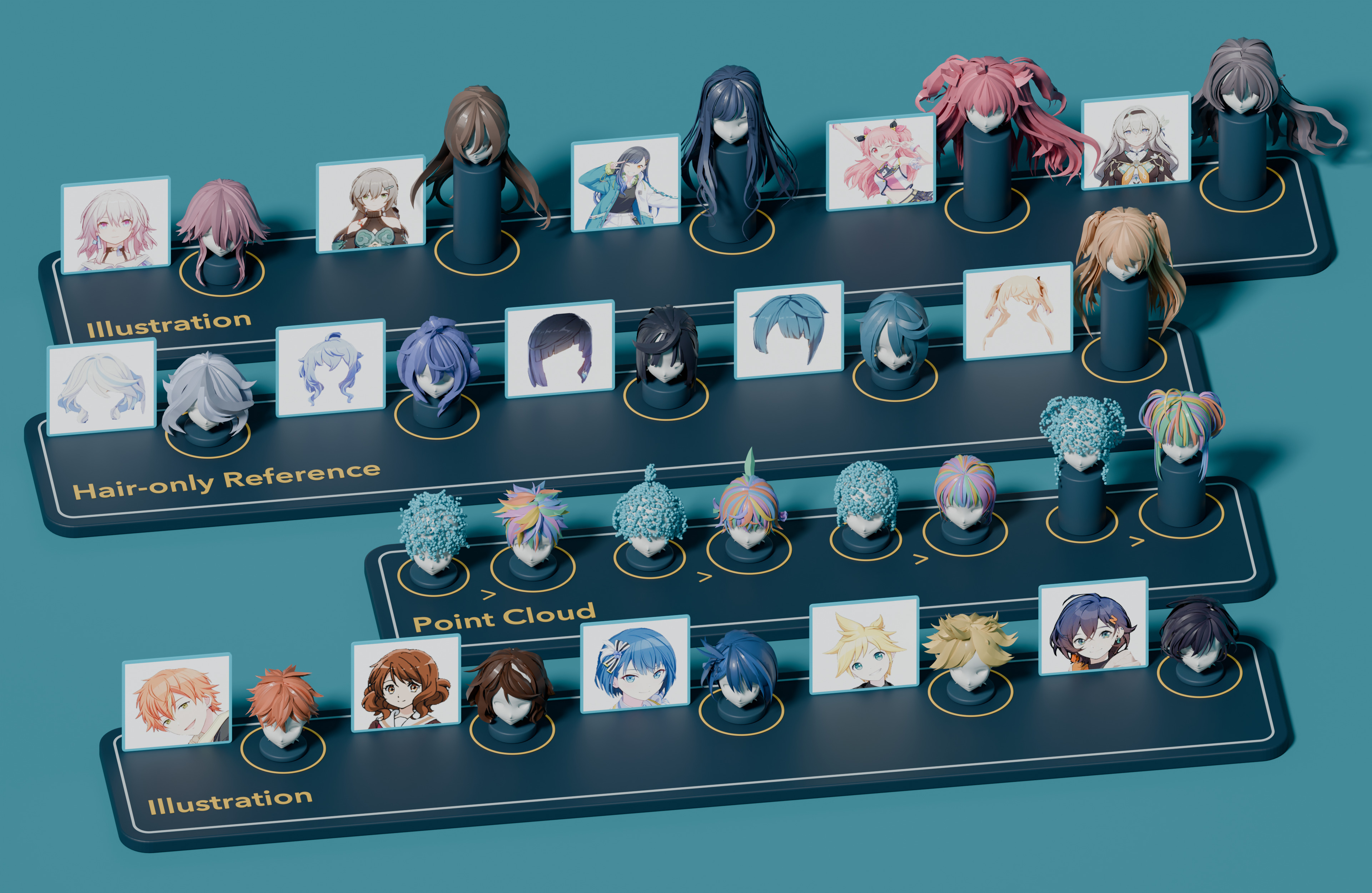}
\captionsetup{hypcap=false}
\captionof{figure}{In-the-wild illustrations, hair-only references and test point clouds (blue), as labeled on the platforms, each to the left of its generated hairstyle. Hair colors are taken from the images for display; point-cloud results show one color per card.}
\captionsetup{hypcap=true}
\label{fig:supp_diorama}%
\end{strip}

\section{Implementation details}
\label{sec:supp_impl}

\paragraph{Card VAE.}
Cards are padded to 128 control points, with the padding masked in attention.
The decoder embeds each arc-length fraction with Fourier features~\cite{tancik2020fourier} of 6 frequencies.
The cross-entropy on the number of control points is weighted by 0.1 and the KL term~\cite{kingma2014auto} by $10^{-6}$.
We train it for 80K steps with batch size 1,024, drawing each batch from cards of similar length, with AdamW~\cite{loshchilov2019decoupled} ($\beta = (0.9, 0.99)$, no weight decay) at learning rate $3 \times 10^{-4}$, a linear warmup over 500 steps, a cosine decay to 10\% of the peak and gradient clipping at norm 1.
It has 8.07M parameters.

\paragraph{Card tokens.}
Root positions and latents are standardized with their mean and standard deviation over the training set; root normals are unit vectors.
Hairstyles with more than 192 cards (0.6\% of the training set) are left out of training.

\paragraph{Set DiT.}
Timesteps are sampled as $t = \operatorname{sigmoid}(\xi)$ with $\xi \sim \mathcal{N}(0, 1)$~\cite{esser2024scaling}.
Both models use AdamW ($\beta = (0.9, 0.99)$, no weight decay) with gradient clipping at norm 1; the learning rate warms up linearly over 2,000 steps for the point-cloud model and 1,000 for the image model and then stays constant; we sample with an exponential moving average (decay 0.9995).
With probability 0.1, learned null tokens replace the condition tokens.
Classifier-free guidance~\cite{ho2022classifier} combines the two velocities as $\mathbf{v}_\varnothing + \gamma\,(\mathbf{v}_{\mathbf{C}} - \mathbf{v}_\varnothing)$ with guidance scale $\gamma = 2$, and the Euler method takes 50 uniform steps from $t = 1$ to $t = 0$.

\paragraph{Point-cloud condition.}
We use the released Dora-VAE 1.1 encoder~\cite{chen2025dora} and cache, for every training point cloud, its 512 output tokens of width 768, taken after the final layer norm and before the KL bottleneck.

\paragraph{Image condition.}
The crop is the square around the union of the hair and face masks, with an 8\% margin on each side.
During training we jitter it: its scale by a log-normal factor ($\sigma = 0.08$), its center by $\mathcal{N}(0, 0.04^2)$ times the side length, and with probability 0.15 we cut off its bottom, keeping 45\% to 80\% of the hair height as in half-body images.
DINOv3~\cite{simeoni2026dinov3} runs at 448 pixels and returns one class token, four register tokens and 784 patch tokens; a layer norm, a linear projection to the model width and a learned positional embedding turn them into the condition.
For in-the-wild images, SAM~3~\cite{carion2026sam3} segments with the prompts ``person'', ``anime character'', ``hair'' and ``face''; the union of the four masks, cleaned with a morphological closing and hole filling, is the matte that cuts the character out onto a white background, and the hair and face masks give the crop.

\paragraph{Head frame.}
\Cref{fig:supp_headframe} illustrates the head frame of \cref{sec:image}.
$W = 0.1276$ and $O = (0, 0.4791, 0.0258)$ are medians over the training set in the dataset frame.
The face width $w$ of a character is the width of its face bounding box, and its anchor is the top center of that box (the center in $x$ and $z$, the top in $y$).
Over the training set, $k$ ranges from 0.70 to 1.24 (5th to 95th percentile).

\paragraph{Training renders.}
Each training character is rendered from 8 views with Blender~4.2~\cite{blender} (EEVEE) at 768 pixels.
For each view we draw a body pose (natural 70\%, hands near the head 25\%, T or A pose 5\%, with jittered IK targets, randomly mirrored one-hand poses, random spine, neck and head rotations and finger curls), a hair state from the VRM 0.x spring-bone simulation~\cite{vrmspec0} (at rest in 35\% of the views, otherwise swinging or settling after a head turn, nod, body turn, jump or sway; in 30\% of the views the hair gravity is set to 0.1 to 0.6), and a camera direction relative to the head (front and three-quarter views 80\%, side 10\%, back 10\%; elevation drawn from $\mathcal{N}(5\text{\textdegree}, (12\text{\textdegree})^2)$ and truncated to $[-25\text{\textdegree}, 40\text{\textdegree}]$).
Lighting is a preset or a random sun, and 80\% of the views draw MToon~\cite{mtoon} outlines.

\paragraph{Appearance augmentation.}
The synthetic style augmentation draws from grayscale, halftone, XDoG~\cite{winnemoller2012xdog} line drawings, full hue rotation, channel permutation, flat colors, painterly filters, thick ink lines, rim light, low resolution, blur, sharpening, noise and JPEG compression.
40\% of the samples keep a light augmentation: brightness, contrast, saturation and hue jitter, blur and JPEG compression with probability 0.3 each, and occasional grayscale.

\paragraph{Restyled images.}
We restyle training renders with Qwen-Image-Edit-2511~\cite{wu2025qwenimage} (20 steps) into gacha (mobile-game) art, thick paint and semi-realistic painting, and keep a restyled image only if its hair and character silhouettes (SAM~3) match the original (IoU at least 0.85 for the hair and 0.9 for the character).
The pool holds 21,336 images, and a training sample is drawn from it with probability 0.33.

\paragraph{Spring-bone chains.}
For the point-cloud model, whose cards are in the dataset frame, we map the generated cards into the head frame with the test character's $k$ and $\mathbf{b}$ and re-encode them with the card VAE before running the chain DiT.
Chains come from the VRM files of the training characters: 32,979 of the 36,440 training hairstyles have hair chains, and leaving out those with more than 48 chains or more than 192 cards leaves 32,599 hairstyles with 438,672 chains.
The chain DiT has 6 attention heads. It is trained for 80K steps with batch size 128, AdamW at learning rate $2 \times 10^{-4}$ after a linear warmup over 1,000 steps, condition dropout 0.1 and an exponential moving average with decay 0.999; all loss weights are 1 except 4 for the existence flag, and the chain slots are randomly permuted for every training sample.
We tune the two skinning thresholds of \cref{sec:chains}, the assignment distance and the ramp length, on 200 random training characters with chains.
For each character, its own chains with their own physics parameters drive 16 points per card under four head motions (a turn, a nod, a jump and a sideways sway), once with the artists' skin weights, where each point takes the weights of the nearest vertex of the hair mesh, and once with our skinning.
We choose the thresholds that maximize the motion correlation of \cref{tab:spring}: sweeping the assignment distance over 0.15, 0.25, 0.4 and 0.6 face widths and the ramp length over 0.01 to 0.3 face widths gives 0.4 and 0.05 (correlation 0.84), and the ramp length barely matters.
On the 58 test hairstyles with chains, our skinning reaches a motion correlation of 0.77 with the artists' skin weights.
The simulation follows the reference spring-bone update of UniVRM~\cite{univrm}: at every frame, the tip of each bone keeps $(1 - \text{drag})$ of its previous displacement, moves toward its rest direction, rotated by the parent, by stiffness times the time step and along gravity by gravity times the time step, and is then pulled back to the bone length.
We use no colliders and run at 60\,fps.
Generated chains use their predicted stiffness, drag 0.4 and no gravity.

\paragraph{Chain metrics.}
In \cref{tab:spring}, lengths are in face widths $W$, and each chain is represented by 8 points along its path.
CD is the Chamfer distance between the path points of the generated and the artists' chains.
Recall and precision are the shares of artist and generated chains whose points lie, on average, within 0.1\,$W$ of the path of a chain in the other set.
For motion, the artists' cards are skinned to each set of chains as in \cref{sec:chains}, the artists' chains keep their own physics parameters, and each rig is simulated under four head motions of 2.5\,s each: a turn (a yaw of 50\textdegree{} in 0.35\,s, then held), a nod (a pitch of 25\textdegree{} down and back in 0.5\,s), a jump (0.15\,m up and down in 0.6\,s) and a sideways sway (6\,cm at 1.5\,Hz for two cycles).
The displacement of each of 16 points per card is taken relative to the moving head, minus its displacement with the head at rest, so that sagging under gravity does not count.
Motion is the correlation over card points between the mean displacement of each point and that under the artists' chains, averaged over the four motions, and Ampl.\ is the mean displacement.
Fit is the share of chain path points within 0.1\,$W$ of the card meshes.

\paragraph{Export.}
Every card mesh carries UV coordinates: the first runs across the width, so the two faces of a card mirror each other and need no seam, and the second is the arc-length fraction, from the root (0) to the tip (1).
To export to VRM, we replace the hair of a character model, its hair bones and its hair spring-bone groups with the generated cards, one spring-bone group per generated chain, and use the geometric skinning of \cref{sec:chains} as skin weights.
Results from in-the-wild images are exported the same way, on the body of a test character with the hair color taken from the image.

\paragraph{Compute.}
Each model trains on one NVIDIA RTX A6000 GPU (48\,GB): the card VAE in 4 hours, the point-cloud model in about 10 hours, the chain DiT in about 3 hours and the image model, starting from the point-cloud model, in about 27 hours.

\section{Additional comparisons and ablations}
\label{sec:supp_tables}

\begin{figure*}[p]
\centering
\includegraphics[width=\linewidth]{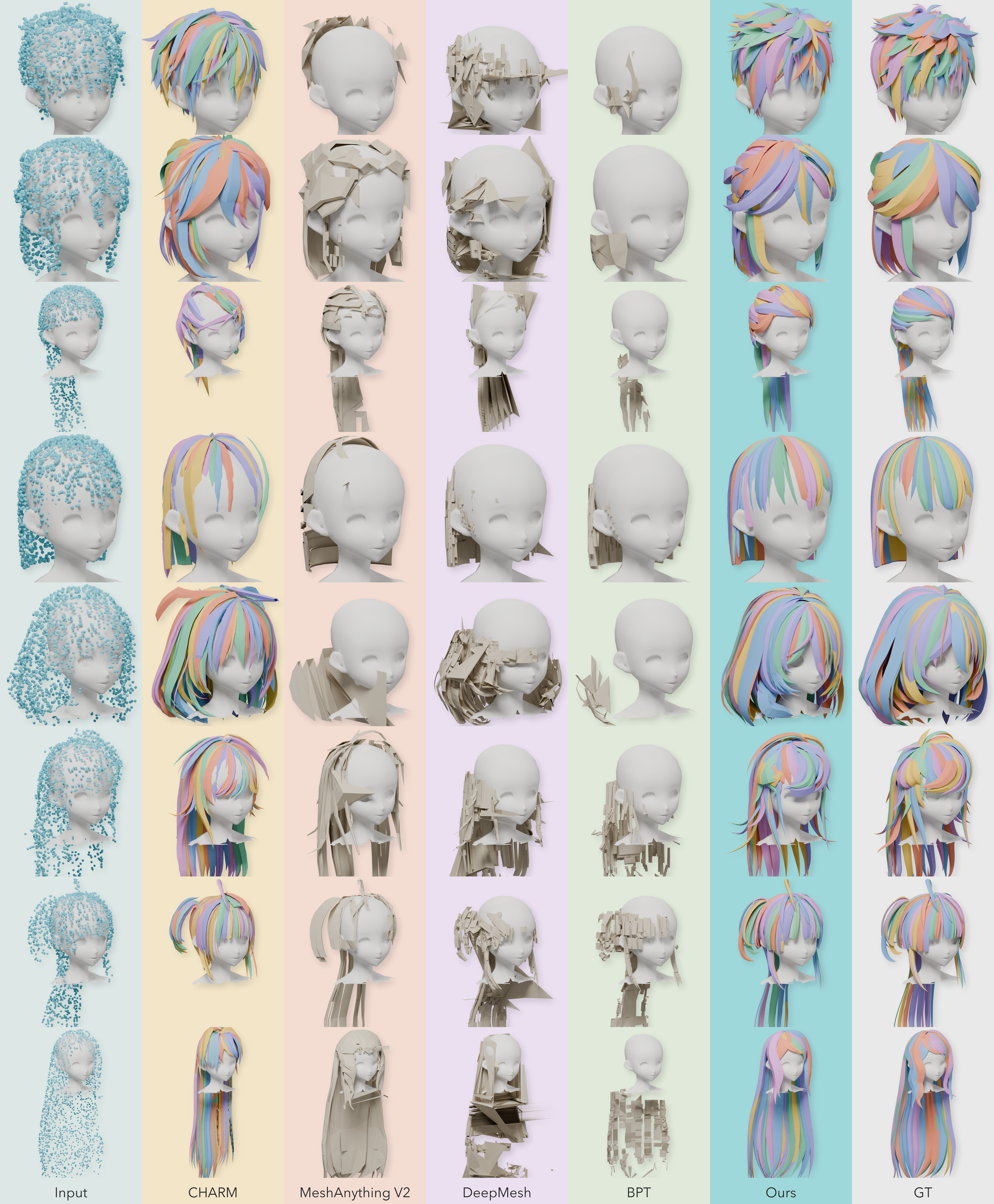}
\caption{More point clouds from CHARM's test set, shown as in \cref{fig:cmp_pc}.}
\label{fig:supp_cmp_pc}
\end{figure*}

\begin{figure*}[p]
\centering
\includegraphics[width=\linewidth]{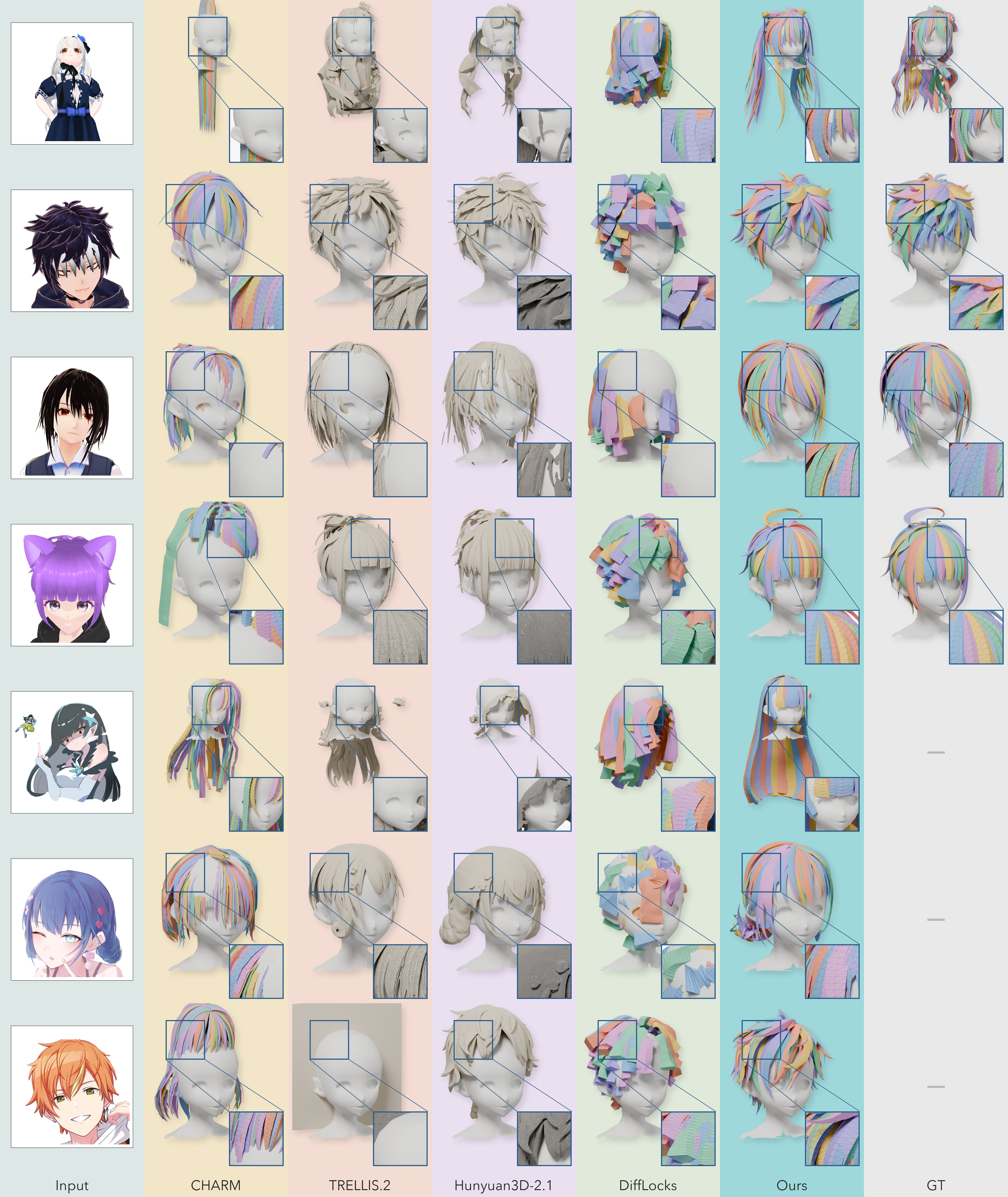}
\caption{Single images, shown as in \cref{fig:cmp_img} with a ground-truth column: four test characters (top) and three in-the-wild illustrations (bottom). TRELLIS.2~\cite{xiang2026trellis2} outputs a flat plate facing the camera for the last one.}
\label{fig:supp_cmp_img1}
\end{figure*}

\begin{figure*}[p]
\centering
\includegraphics[width=\linewidth]{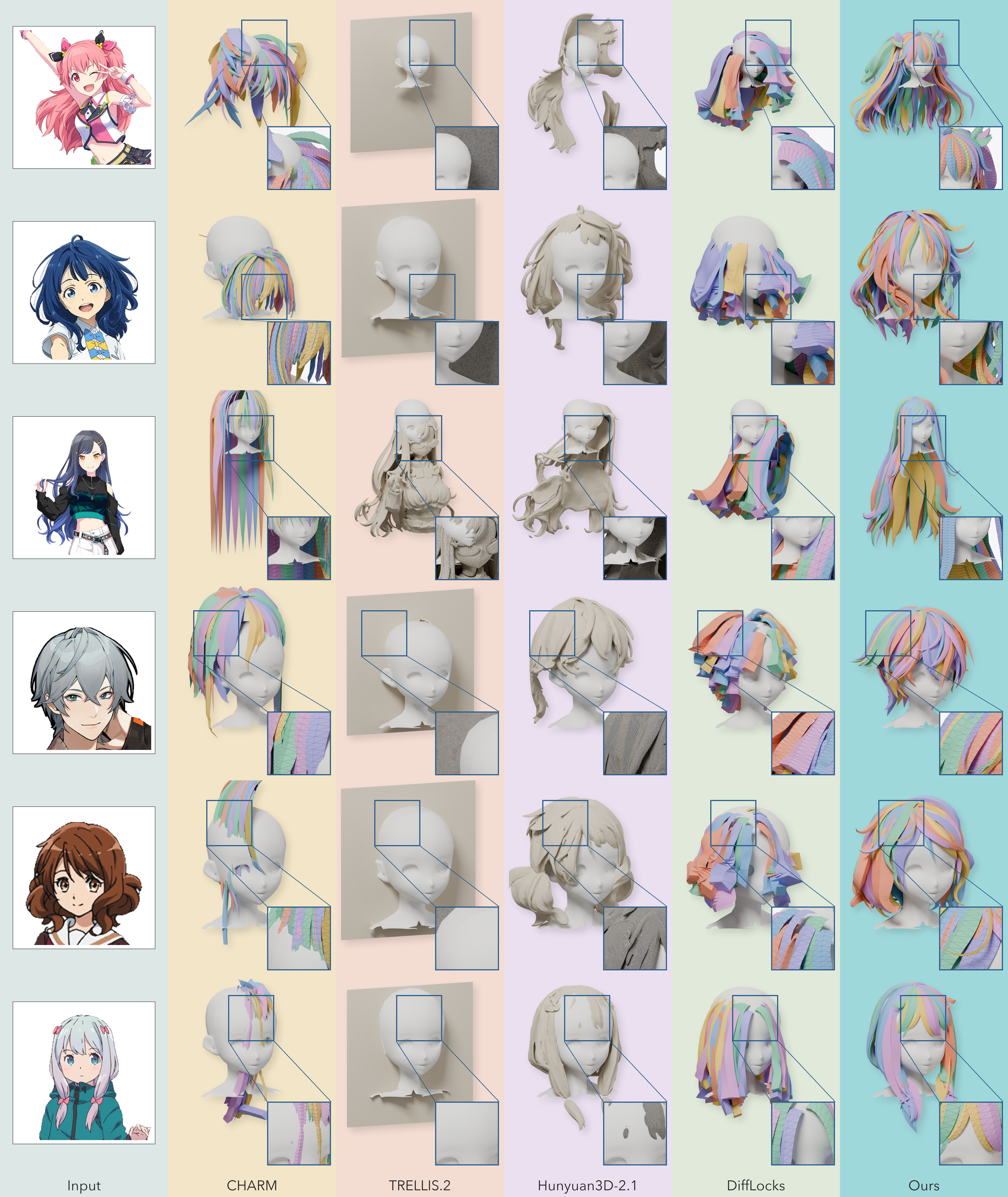}
\caption{Six more in-the-wild illustrations, shown as in \cref{fig:cmp_img}. TRELLIS.2 outputs a flat plate facing the camera for five of them.}
\label{fig:supp_cmp_img2}
\end{figure*}

\begin{figure*}[t]
\centering
\includegraphics[width=\linewidth]{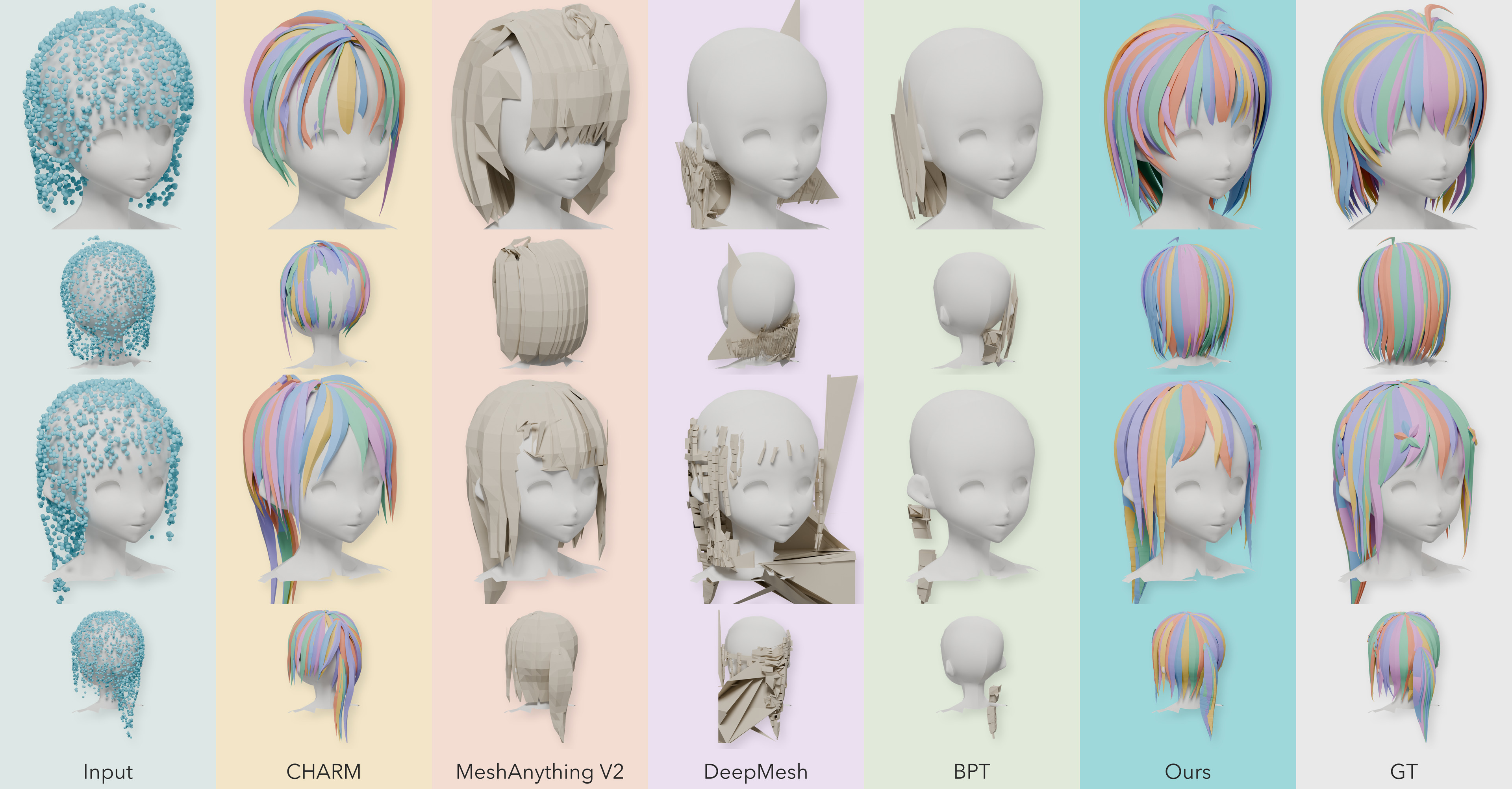}
\caption{Two more point clouds from CHARM's test set, shown as in \cref{fig:cmp_pc}, each with a row of back views below it.}
\label{fig:supp_cmp_pc_back}
\end{figure*}

\begin{table}[t]
\centering
\caption{Cascades of image-to-3D generators (TRELLIS.2~\cite{xiang2026trellis2} and Hunyuan3D-2.1~\cite{yang2025hunyuan3d21}) and point-cloud generators, against \method. ``$\rightarrow$'' feeds an image-to-3D output, sampled as a point cloud, to a point-cloud generator; PC model: our point-cloud model. Columns as in \cref{tab:img}: aligned CD and IoU, failure rates (\cref{sec:setup}) and time per image.}
\label{tab:supp_cascade}
\footnotesize
\setlength{\tabcolsep}{1.9pt}
\begin{tabular}{@{}lccccc@{}}
\toprule
& \multicolumn{2}{c}{Test characters} & \multicolumn{2}{c}{Failure rate$\downarrow$} & \\
\cmidrule(lr){2-3}\cmidrule(lr){4-5}
Method & CD$\downarrow$ & IoU$\uparrow$ & Test & In the wild & Time$\downarrow$ \\
\midrule
TRELLIS.2 $\rightarrow$ CHARM & 0.0170 & 0.554 & 32.9\% & 80.0\% & 119\,s \\
Hunyuan3D-2.1 $\rightarrow$ CHARM & 0.0163 & 0.573 & 28.9\% & 80.0\% & 96.8\,s \\
TRELLIS.2 $\rightarrow$ PC model & \underline{0.0138} & \underline{0.621} & \underline{22.4\%} & \underline{76.7\%} & 85.8\,s \\
Hunyuan3D-2.1 $\rightarrow$ PC model & 0.0143 & 0.615 & 29.4\% & 80.0\% & \underline{64.2}\,s \\
\midrule
\method & \textbf{0.0089} & \textbf{0.725} & \textbf{3.5\%} & \textbf{10.0\%} & \textbf{1.33}\,s \\
\bottomrule
\end{tabular}
\end{table}

\begin{table}[t]
\centering
\caption{All ablations on point clouds; each row changes one part of \method (rows of \cref{tab:abl_pc}: \cref{sec:ablations}; card tokens: \cref{sec:supp_orient}). CD and IoU on CHARM's ground truth and Orient. on the original meshes, as in \cref{tab:pc}; turn: the largest angle between consecutive segments of a card, median per hairstyle (artist cards: 14.1\textdegree).
Root position inside the latent: the card VAE encodes absolute positions, and the token has no separate root position. Bold: best in each column with an arrow.}
\label{tab:supp_abl_pc}
\footnotesize
\setlength{\tabcolsep}{3.5pt}
\begin{tabular}{@{}lcccc@{}}
\toprule
Variant & CD$\downarrow$ & IoU$\uparrow$ & Orient.$\downarrow$ & Turn \\
\midrule
\method & \textbf{0.0066} & \textbf{0.867} & 16.8\textdegree & 14.2\textdegree \\
\midrule
\multicolumn{5}{@{}l}{\emph{Card token}} \\
\quad raw control points & 0.0068 & 0.856 & 17.1\textdegree & 21.9\textdegree \\
\quad PCA, 64 coefficients & 0.0078 & 0.832 & 17.8\textdegree & 11.1\textdegree \\
\quad VAE with index decoding & 0.0079 & 0.823 & 18.1\textdegree & 25.3\textdegree \\
\quad latent $\mathbf{z} \in \mathbb{R}^{64}$ & 0.0071 & 0.843 & 17.2\textdegree & 13.4\textdegree \\
\multicolumn{5}{@{}l}{\emph{Token composition}} \\
\quad w/o root normal & 0.0070 & 0.850 & 26.4\textdegree & 13.9\textdegree \\
\quad w/o existence (card-count head) & 0.0068 & 0.859 & \textbf{16.7}\textdegree & 14.5\textdegree \\
\quad root position inside the latent & 0.0067 & 0.859 & 17.2\textdegree & 14.6\textdegree \\
\multicolumn{5}{@{}l}{\emph{Data, encoder and size}} \\
\quad CHARM's card templates & 0.0068 & 0.856 & 26.7\textdegree & 15.8\textdegree \\
\quad CHARM's encoder~\cite{zhao2023michelangelo} & 0.0068 & 0.853 & 17.2\textdegree & 14.0\textdegree \\
\quad small model (54.6M) & 0.0072 & 0.847 & 17.5\textdegree & 14.2\textdegree \\
\bottomrule
\end{tabular}
\end{table}

\begin{table*}[t]
\centering
\caption{Ablations of the image model on the 76 test characters. Front view: CD before alignment (CHARM's protocol, so it includes where the hair sits), overlap (IoU) of the reprojected hair with the input's hair, and aligned CD as in \cref{tab:img}. Restyled: the front views restyled as in \cref{fig:supp_style}, with aligned CD, root precision P and recall R as in \cref{sec:setup}, and cards per hairstyle (ground truth: 65.1). Style aug.: the synthetic style augmentation of \cref{sec:supp_impl}; without it, every sample gets only the light augmentation.
Bold: best in each column with an arrow.}
\label{tab:supp_abl_img}
\footnotesize
\setlength{\tabcolsep}{3pt}
\begin{tabular}{@{}lccccccc@{}}
\toprule
& \multicolumn{3}{c}{Front view} & \multicolumn{4}{c}{Restyled} \\
\cmidrule(lr){2-4}\cmidrule(lr){5-8}
Variant & CD$\downarrow$ & Overlap$\uparrow$ & Aligned CD$\downarrow$ & Aligned CD$\downarrow$ & P$\uparrow$ & R & Cards \\
\midrule
\method & 0.0136 & \textbf{0.829} & \textbf{0.0089} & 0.0096 & \textbf{0.849} & 0.833 & 68.9 \\
\quad w/o head frame & 0.0189 & 0.798 & \textbf{0.0089} & \textbf{0.0095} & 0.736 & 0.756 & 69.8 \\
\quad w/o restyled images & \textbf{0.0135} & 0.827 & 0.0090 & 0.0097 & 0.781 & 0.826 & 86.9 \\
\quad w/o restyled images and style aug. & 0.0137 & 0.827 & 0.0091 & 0.0100 & 0.760 & 0.790 & 79.9 \\
\bottomrule
\end{tabular}
\end{table*}

\begin{figure*}[t]
\centering
\includegraphics[width=\linewidth]{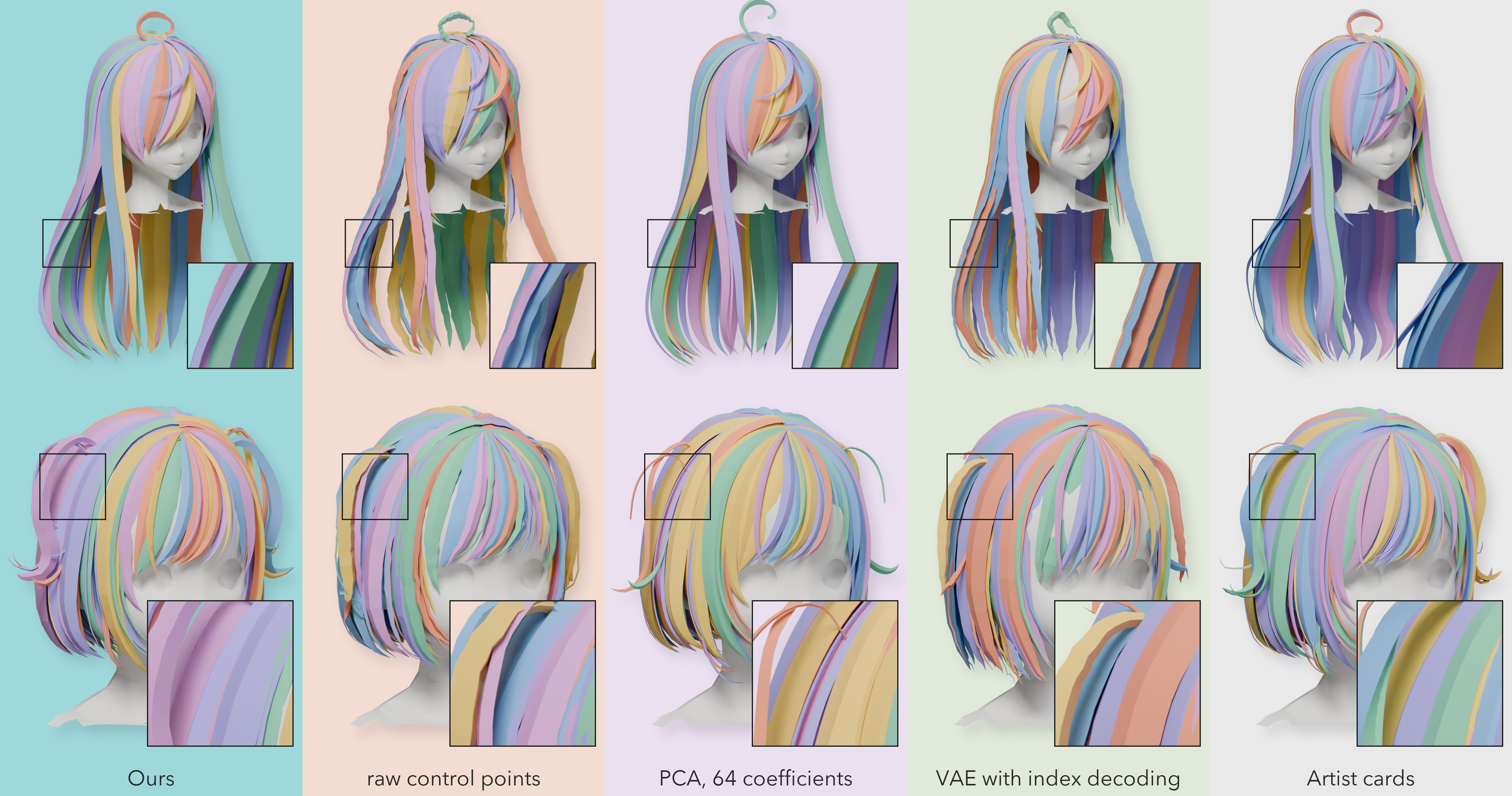}
\caption{Card tokens. A set DiT of the same architecture, trained with each card token of \cref{tab:supp_abl_pc}, generates two test hairstyles from point clouds; insets magnify the same window in each row. Raw control points and the VAE with index decoding give jagged cards, PCA with 64 coefficients gives straighter cards than the artists', and our latent gives cards as smooth as theirs. Each card has its own color.}
\label{fig:supp_cmp_repr}
\end{figure*}

\begin{table}[t]
\centering
\caption{Card VAE design: reconstruction of the cards of CHARM's 100 test hairstyles, with $\mathbf{z} \in \mathbb{R}^{32}$. Per card, CD is the two-sided point-to-surface distance between the reconstructed and the input card, both meshed with the input card's root normal, in $10^{-3}$ of the character height, and F1 is the F-score of their surface samples at distance 0.001; exact $n$: share of cards with the right number of control points; turn as in \cref{tab:supp_abl_pc} (artist cards: 14.1\textdegree). +: added to arc-length decoding, where $\mathbf{z}$ is added to every decoder query; uniform targets: training queries the decoder at $n$ uniformly spaced arc-length fractions, as at test time. Bold: best in each column with an arrow.}
\label{tab:supp_vae}
\footnotesize
\setlength{\tabcolsep}{2.4pt}
\begin{tabular}{@{}lcccc@{}}
\toprule
Card code & CD ($10^{-3}$)$\downarrow$ & F1$\uparrow$ & Exact $n$$\uparrow$ & Turn \\
\midrule
index decoding & 4.17 & 0.411 & 0.917 & 25.3\textdegree \\
arc-length decoding & 0.98 & 0.893 & \textbf{0.984} & 15.6\textdegree \\
\quad + adaLN & 0.91 & 0.904 & \textbf{0.984} & 15.8\textdegree \\
\quad + world-unit loss & 0.84 & 0.920 & \textbf{0.984} & 13.1\textdegree \\
\quad + both & \textbf{0.70} & 0.942 & \textbf{0.984} & 13.7\textdegree \\
\quad + both + uniform targets (ours) & \textbf{0.70} & \textbf{0.943} & \textbf{0.984} & 14.3\textdegree \\
\bottomrule
\end{tabular}
\end{table}

\begin{table}[t]
\centering
\caption{Width direction and shape of artist cards rebuilt from their control points, on CHARM's 100 test hairstyles: control points as placed by the artists or quantized by CHARM, the root direction from the first 3 points (their smallest principal axis, as in CHARM's templates) or from the original mesh, and CHARM's least squares, a Bishop frame or a constant direction along the card. Mean: orientation error as in \cref{tab:pc}; median: the same with the median over each hairstyle's cards; P2S: two-sided point-to-surface distance to the original card mesh, with CHARM's widths and thicknesses in every row so that only the width direction differs (card VAE: its own). Bold: best above the line.}
\label{tab:supp_orient_gt}
\footnotesize
\setlength{\tabcolsep}{2.4pt}
\begin{tabular}{@{}llccc@{}}
\toprule
Root direction & Along the card & Mean$\downarrow$ & Median$\downarrow$ & P2S ($10^{-3}$)$\downarrow$ \\
\midrule
\multicolumn{5}{@{}l}{\emph{Control points placed by the artists}} \\
first 3 points & least squares & 24.7\textdegree & 19.7\textdegree & 1.39 \\
first 3 points & Bishop & 16.3\textdegree & 9.4\textdegree & 1.24 \\
original mesh & least squares & 19.3\textdegree & 16.9\textdegree & 1.36 \\
original mesh & constant & 11.9\textdegree & 9.7\textdegree & 1.28 \\
original mesh & Bishop & \textbf{7.5}\textdegree & \textbf{5.6}\textdegree & \textbf{1.10} \\
\multicolumn{5}{@{}l}{\emph{Control points quantized by CHARM}} \\
first 3 points & least squares & 40.6\textdegree & 37.4\textdegree & 2.20 \\
first 3 points & Bishop & 35.3\textdegree & 30.7\textdegree & 2.39 \\
original mesh & least squares & 19.6\textdegree & 17.2\textdegree & 1.43 \\
original mesh & Bishop & 8.2\textdegree & 6.3\textdegree & 1.19 \\
\midrule
\multicolumn{5}{@{}l}{\emph{Card VAE reconstruction}} \\
original mesh & Bishop & 8.3\textdegree & 6.5\textdegree & 1.29 \\
\bottomrule
\end{tabular}
\end{table}

\begin{table}[t]
\centering
\caption{Width direction of generated cards on CHARM's 100 test point clouds. Only the width direction used for meshing changes. Orient.\ and CD: as in \cref{tab:pc}; root: angle at the root, median over the cards of each hairstyle.}
\label{tab:supp_orient_gen}
\footnotesize
\setlength{\tabcolsep}{3pt}
\begin{tabular}{@{}llccc@{}}
\toprule
Method & Width direction & Orient.$\downarrow$ & Root$\downarrow$ & CD$\downarrow$ \\
\midrule
CHARM & first 3 points + least squares & 40.0\textdegree & 33.4\textdegree & 0.0117 \\
CHARM & first 3 points + Bishop & 37.5\textdegree & 33.1\textdegree & 0.0119 \\
\method & first 3 points + least squares & 30.2\textdegree & 19.4\textdegree & 0.0067 \\
\method & root normal + Bishop & \textbf{16.8}\textdegree & \textbf{8.9}\textdegree & \textbf{0.0066} \\
\bottomrule
\end{tabular}
\end{table}

\Cref{fig:supp_cmp_pc} shows eight more hairstyles generated from the point clouds of CHARM's test set, and \cref{fig:supp_cmp_img1,fig:supp_cmp_img2} compare the single-image methods on four test characters and nine in-the-wild illustrations.
\Cref{fig:supp_cmp_pc_back} adds back views on two more point clouds: CHARM~\cite{he2025charm} leaves part of the back of the head bare on the first, and DeepMesh~\cite{zhao2025deepmesh} and BPT~\cite{weng2024scaling} cover only parts of the back.
\Cref{tab:supp_cascade} adds cascades of image-to-3D and point-cloud generators to \cref{tab:img}, \cref{tab:supp_abl_pc} lists all point-cloud ablations, and \cref{tab:supp_abl_img} the ablations of the image model.
For the head-frame ablation of \cref{sec:ablations}, we also render every test character from the side, with the pose, hair motion and lighting of its front view and the camera turned by 90\textdegree{} around the head.

\section{Card representation and orientation}
\label{sec:supp_orient}
\Cref{fig:supp_cmp_repr} compares the card tokens of \cref{tab:supp_abl_pc}.

\paragraph{Card VAE design.}
\Cref{tab:supp_vae} starts from a VAE that decodes the control points by index, as StrokeFusion~\cite{zhou2026strokefusion} decodes the points of a stroke.
It takes the same inputs as our card VAE; its encoder sees the control points without their arc-length fractions, and its decoder adds $\mathbf{z}$ to 128 learned slot embeddings and predicts in each slot a control point and whether the slot holds one, with a squared error on standardized features.
Decoding at arc-length fractions instead cuts the reconstruction error by a factor of 4.3, and adaLN modulation~\cite{Peebles2022DiT} and the world-unit loss of \cref{sec:card_token}, in place of a squared error on standardized features, together lower it by another 29\%.
The raw token resamples a card to 64 points by arc length and holds their positions relative to the root position, widths and thicknesses (317 values) with the number of control points; the PCA token holds the first 64 principal components of these 317 values, fitted on 300,000 training cards and scaled to unit variance.

\paragraph{Width direction of the data.}
Our cards keep the control point positions of CHARM's templates; they differ in the width direction and in the widths and thicknesses, which we measure along the Bishop frame~\cite{bishop1975there}.
\Cref{tab:supp_orient_gt} rebuilds the artist cards with each choice of control points, root direction and propagation, and measures both the width direction and the shape.
CHARM's templates take the root direction from the first three control points and solve the rest of the card by least squares, which keeps each width direction perpendicular to the card and close to its neighbors.
On the control points quantized into CHARM's tokens (512 levels per position axis), least squares rebuilds the shape more closely than a Bishop frame from the same start, as CHARM found for Bishop propagation from estimated normals~\cite{he2025charm}, but both leave the width direction far off (40.6\textdegree{} and 35.3\textdegree), because the quantization turns the root direction from the first three points by 23.6\textdegree{} (median).
Without the quantization, the Bishop frame is closer in both direction and shape, and with the root direction of the original mesh it is closer still, on artist and quantized points alike (7.5\textdegree{} and 8.2\textdegree, against 19.3\textdegree{} and 19.6\textdegree{} for least squares): given the right root direction, the quantization costs less than 1\textdegree.
Our cards take this route, with the root normal of the original mesh propagated by a Bishop frame; the card VAE adds 0.8\textdegree, and keeping the root direction constant along the card gives 11.9\textdegree.
Our data cards are also closer to the original card meshes, with a two-sided point-to-surface distance of $0.945 \times 10^{-3}$ against $1.388 \times 10^{-3}$ for CHARM's templates.

\paragraph{Width direction of the outputs.}
\Cref{tab:supp_orient_gen} shows the same on generated cards.
On CHARM's outputs, a Bishop frame from the first three points lowers the orientation error only from 40.0\textdegree{} to 37.5\textdegree{} and raises the CD slightly (0.0117 to 0.0119), because the root direction stays 33\textdegree{} off.
On our outputs, the generated root normal with a Bishop frame is better in both than least squares from the first three points (16.8\textdegree{} against 30.2\textdegree, CD 0.0066 against 0.0067).

\clearpage

\section{Baseline details}
\label{sec:supp_baselines}
\paragraph{Alignment.}
For the aligned CD and IoU of \cref{tab:img,tab:supp_cascade}, every output is normalized to the bounding box of the ground truth and then aligned to it by ICP with scale, started from 24 axis-aligned rotations.

\paragraph{CHARM.}
We run the authors' checkpoints with greedy decoding: the point-cloud model on the point clouds of CHARM's test set, and the image model on a line drawing of the hair made with CHARM's pipeline.

\paragraph{Card-level autoregression.}
The card-level autoregressive model of \cref{tab:pc}, with 182.1M parameters, uses our card tokens, card VAE, data, condition and training steps.
A 17-layer transformer of width 768 generates the cards in CHARM's order, by the azimuth of each card's mean point around the vertical axis, and a 6-layer per-token diffusion head, as in MAR~\cite{li2024mar}, samples each token with 50 steps and guidance scale 2; before each card, a stop head decides whether the sequence ends.

\paragraph{Retrieval.}
Retrieval in \cref{tab:pc} compares the mean of the input's condition tokens with those of the 36,440 training hairstyles by cosine similarity and returns the cards of the closest.
For the chains in \cref{tab:spring}, it returns the chains of the closest training hairstyle that has chains.

\paragraph{Procedural rig.}
The procedural rig of \cref{tab:spring} places chains on cards by rules, with parameters from the spring bones of 300 training characters and lengths in face widths $W$.
Cards whose centerlines lie close (mean distance below 0.2\,$W$) and point the same way form a unit, which gets one chain per 0.23\,$W$ of its width.
A chain follows the mean centerline of its cards, fitted by a quadratic B\'ezier curve, and starts 15\% of the card length from the root.
Longer chains get more joints and a lower stiffness, as in the training characters; all chains use drag 0.4 and no gravity.

\paragraph{Mesh generators.}
MeshAnything V2~\cite{chen2024meshanythingv2}, DeepMesh and BPT generate a mesh from a point cloud of the ground-truth hair mesh, with their official checkpoints and settings.
MeshAnything V2 and BPT sample it from the marching-cubes surface of the hair, and DeepMesh from the mesh itself.
The official sampler of DeepMesh mirrors its output, which we undo.
We sample 10K points from depth renders of each output over 100 views, as CHARM scans its data, and evaluate them with CHARM's own scripts.

\paragraph{Image-to-3D generators and cascades.}
TRELLIS.2 and Hunyuan3D-2.1 generate geometry only, from the image with all but the hair masked out.
For the cascades of \cref{tab:supp_cascade}, the mesh is placed in the median hair bounding box of the training set, scanned into 10K points as above and passed to CHARM or our point-cloud model.

\paragraph{DiffLocks.}
We run the official checkpoints of DiffLocks~\cite{rosu2025difflocks}.
Its pipeline crops the image around a face box, which we take from our face mask on every input.
Each output holds about 95K strands, which we convert into $K = 60$ cards: every strand is resampled to 16 points by arc length, k-means groups the strands into $K$ clusters, and each cluster becomes one card whose centerline is the mean of its strands and whose width, thickness and width direction come from the spread of the strands around it.
Strands laid inside the ground-truth cards and converted back reach an aligned CD of 0.0050, and $K = 40$, 100 and 150 give 0.0172, 0.0174 and 0.0172, against 0.0174 for $K = 60$.

\section{Additional results}
\label{sec:supp_results}

\begin{figure*}[p]
\centering
\includegraphics[height=8.3in]{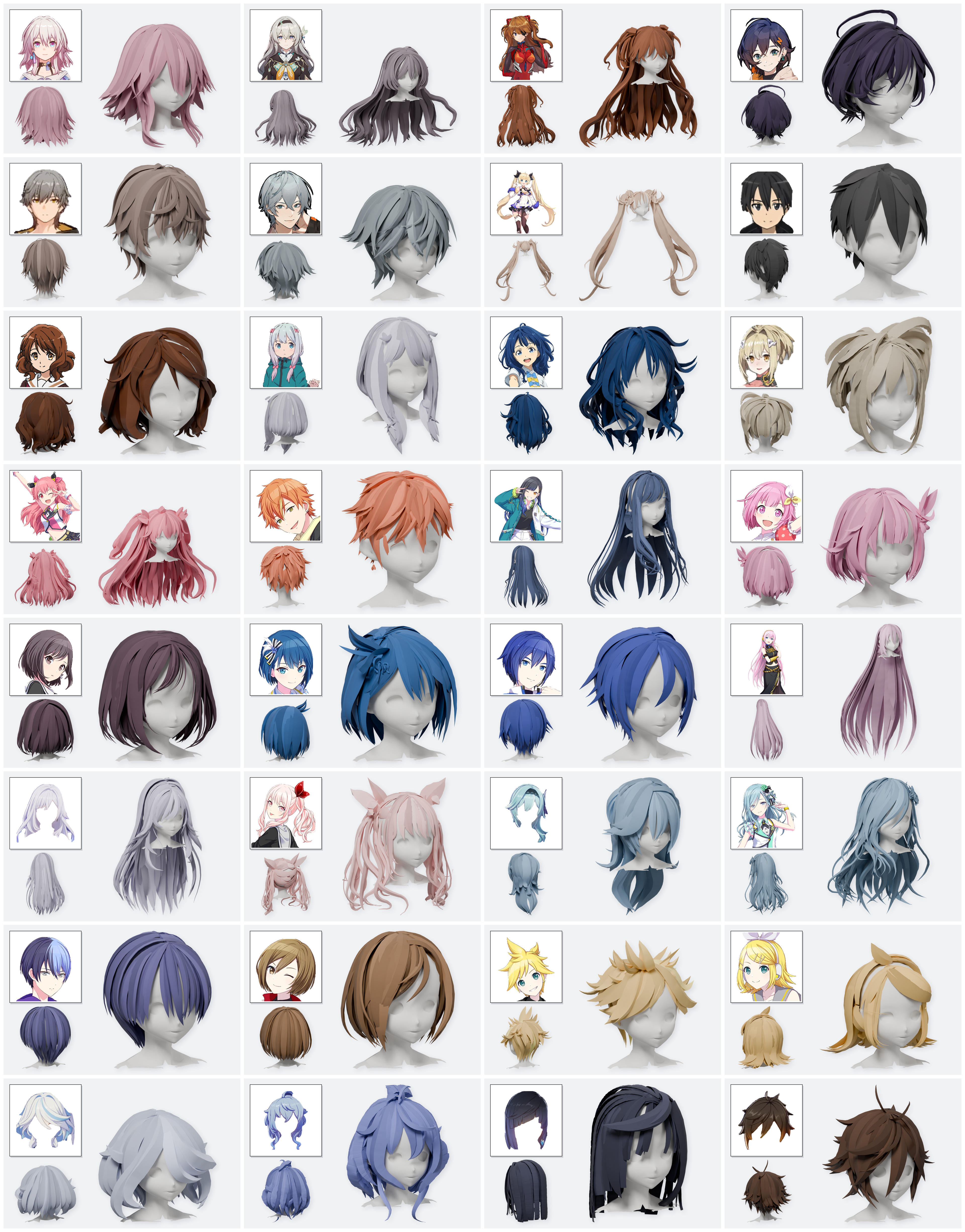}
\caption{More hairstyles generated from in-the-wild images. Each cell shows the input (top left), the back view (bottom left) and the front view of the generated hairstyle. Hair colors are taken from the images for display.}
\label{fig:supp_gallery_uc}
\end{figure*}

\begin{figure*}[p]
\centering
\includegraphics[height=8.3in]{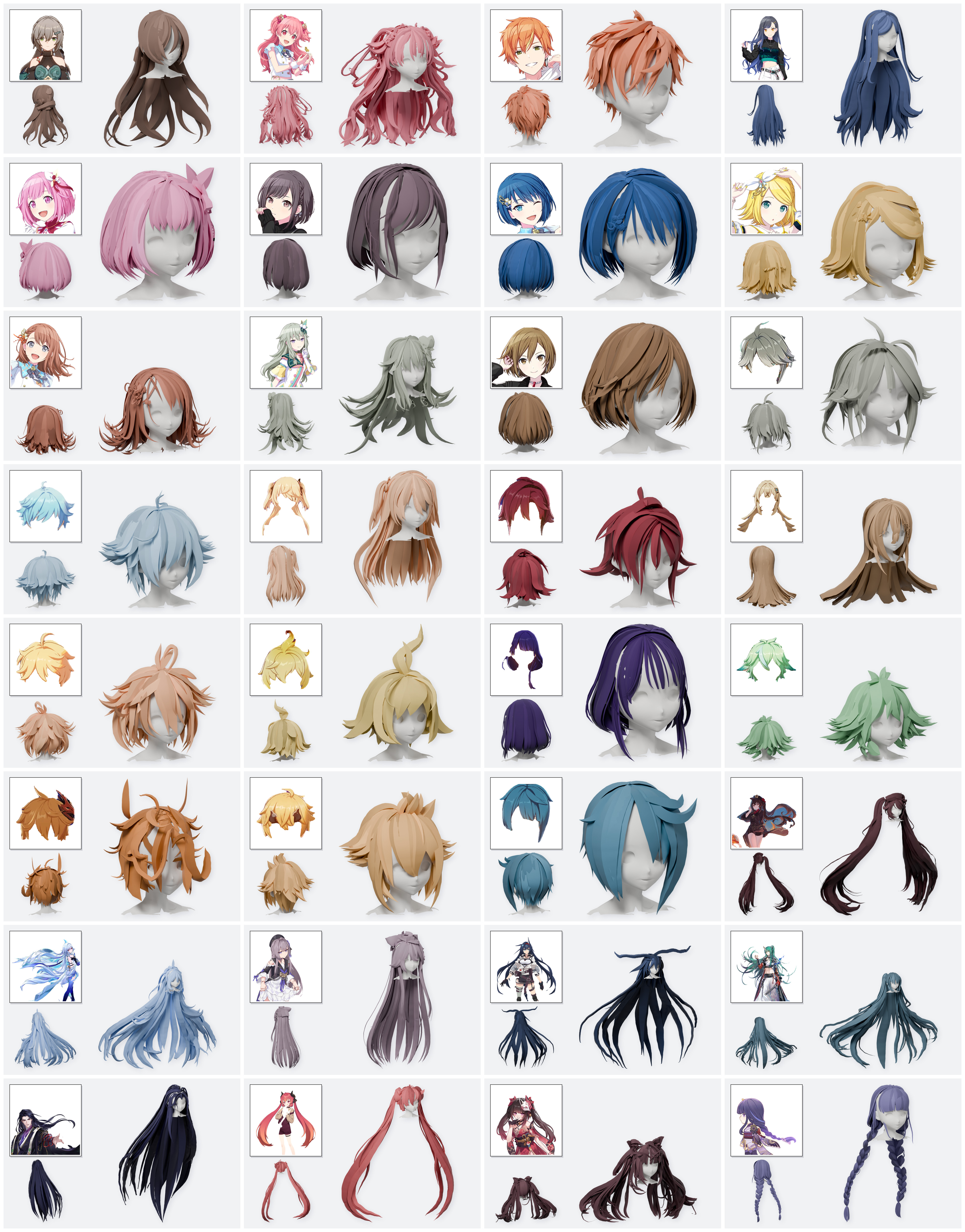}
\caption{More hairstyles generated from in-the-wild images, shown as in \cref{fig:supp_gallery_uc}.}
\label{fig:supp_gallery_uc2}
\end{figure*}

\begin{figure*}[p]
\centering
\includegraphics[width=\linewidth]{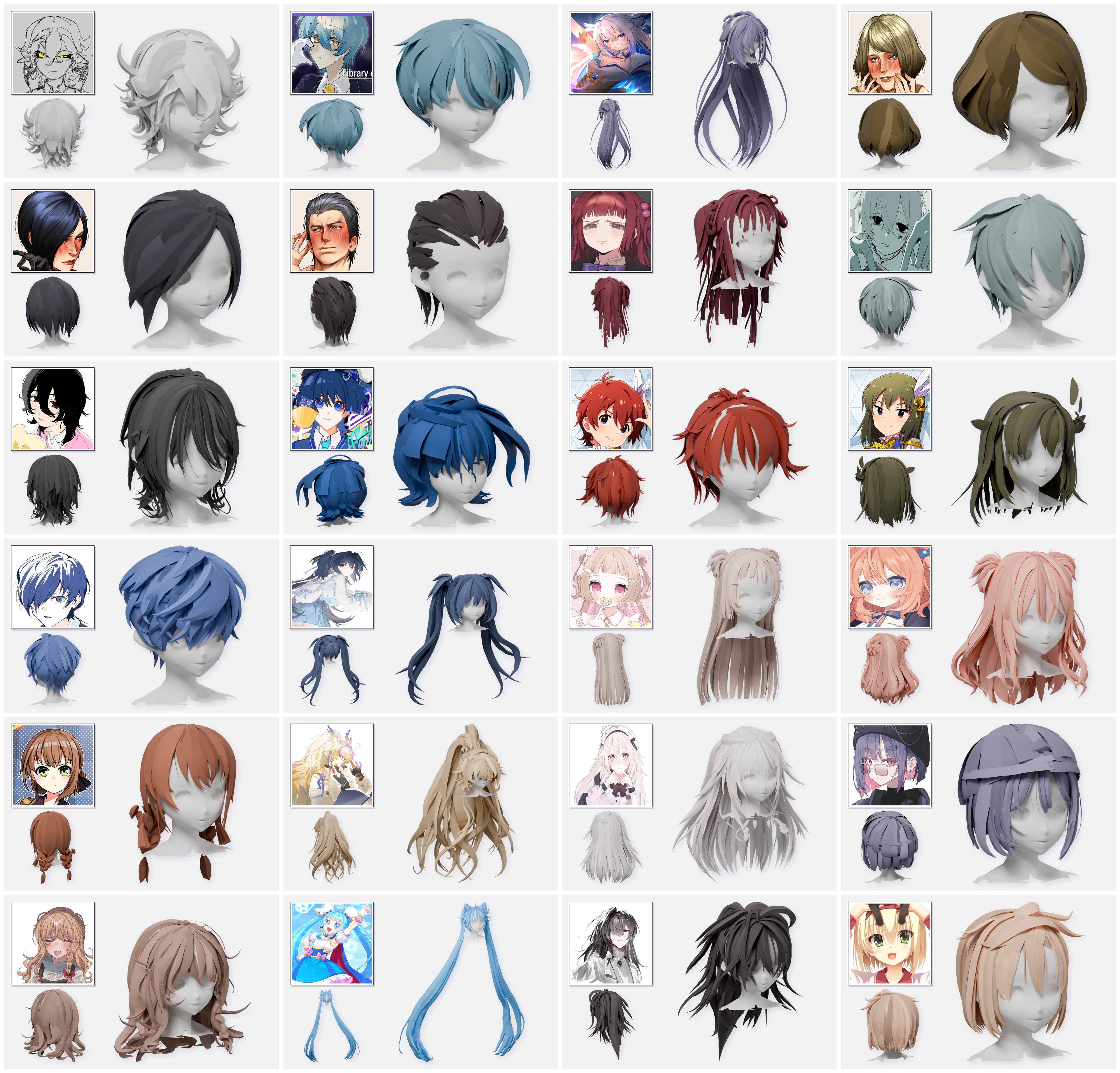}
\caption{More hairstyles generated from in-the-wild images, shown as in \cref{fig:supp_gallery_uc}.}
\label{fig:supp_gallery_illus1}
\end{figure*}

\begin{figure*}[p]
\centering
\includegraphics[width=\linewidth]{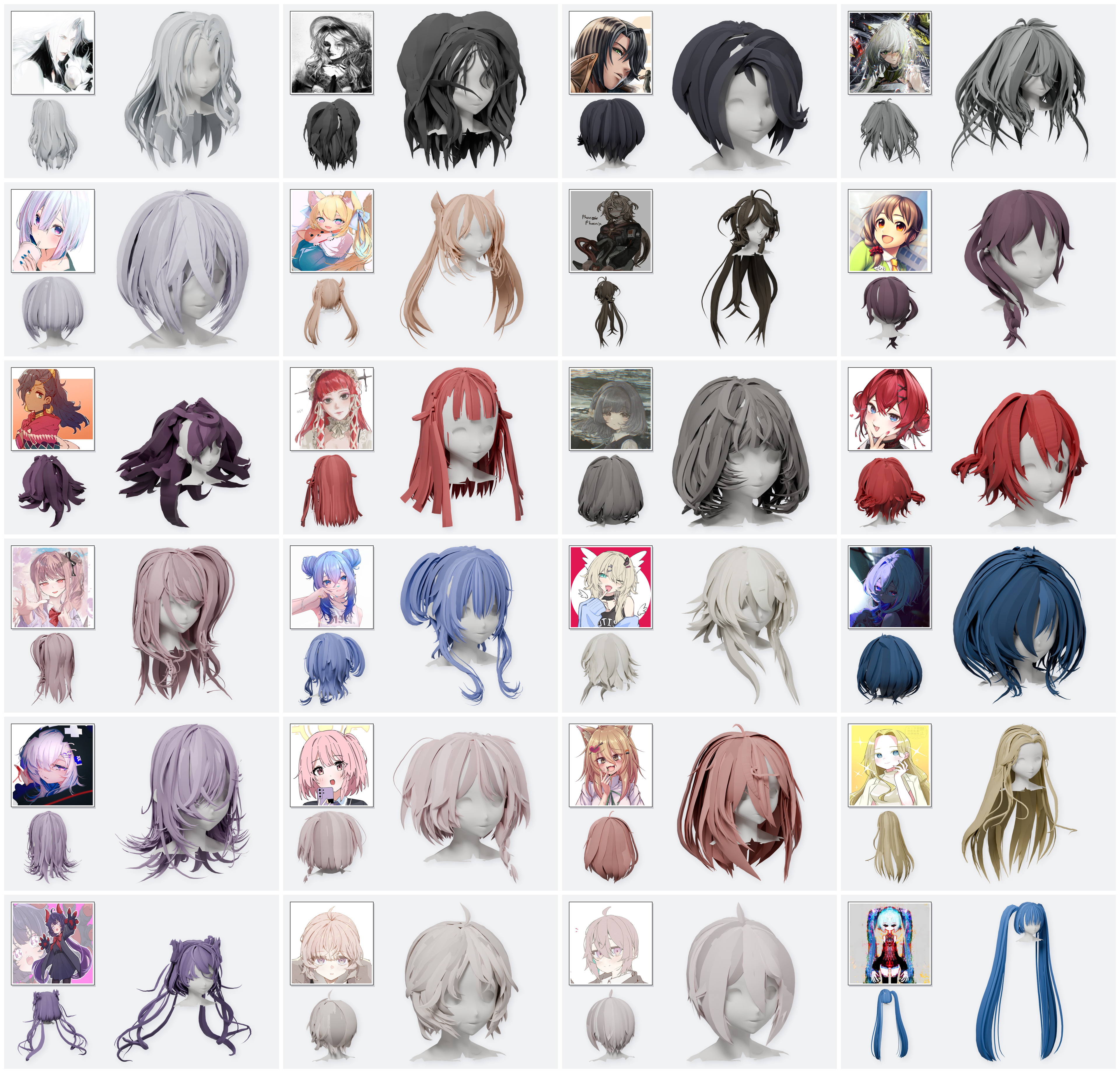}
\caption{More hairstyles generated from in-the-wild images, shown as in \cref{fig:supp_gallery_uc}.}
\label{fig:supp_gallery_illus2}
\end{figure*}

\paragraph{More results.}
\Cref{fig:supp_gallery_uc,fig:supp_gallery_uc2,fig:supp_gallery_illus1,fig:supp_gallery_illus2} show 112 more hairstyles generated from in-the-wild images, and \cref{fig:supp_diorama} places results from illustrations, hair-only references (no character) and point clouds next to their inputs.

\begin{figure*}[t]
\centering
\includegraphics[width=\linewidth]{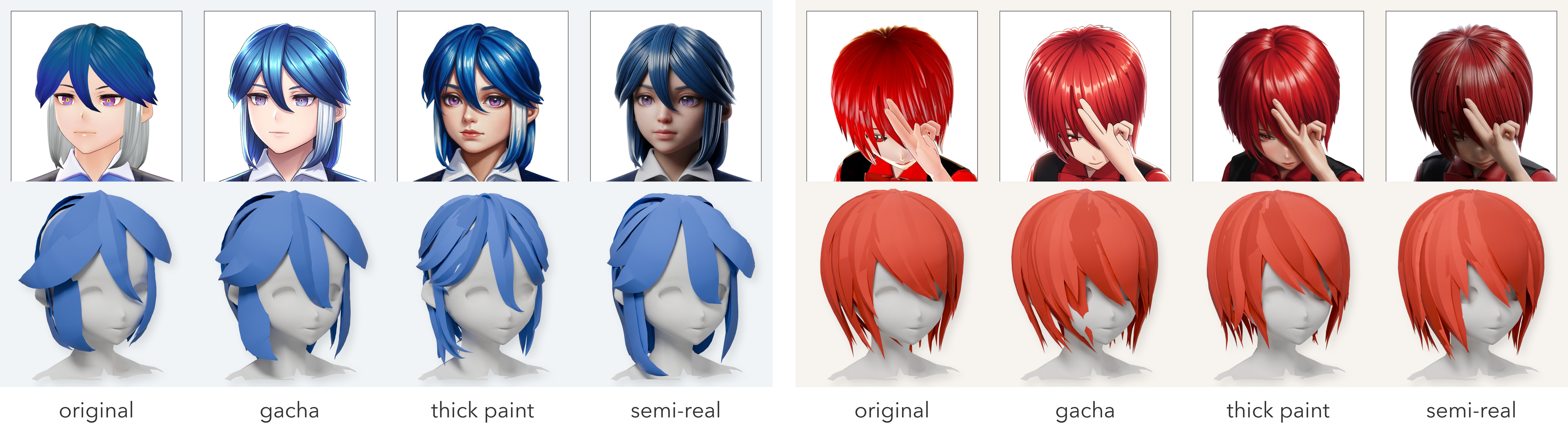}
\caption{Drawing styles. Two test characters, original and restyled by an image-editing model~\cite{wu2025qwenimage} into the gacha, thick paint and semi-real styles (top row of each), and the hairstyle generated from each image (bottom row). Each character's hair is shown in one color.}
\label{fig:supp_style}
\end{figure*}

\paragraph{Drawing styles.}
\Cref{fig:supp_style} shows two test characters restyled into gacha art, thick paint and semi-realistic painting, with the hairstyle generated from each image.
\Cref{tab:supp_style} evaluates every style on all 76 test characters for the three models of the style ablations in \cref{tab:supp_abl_img}.

\begin{figure*}[t]
\centering
\includegraphics[width=0.85\linewidth]{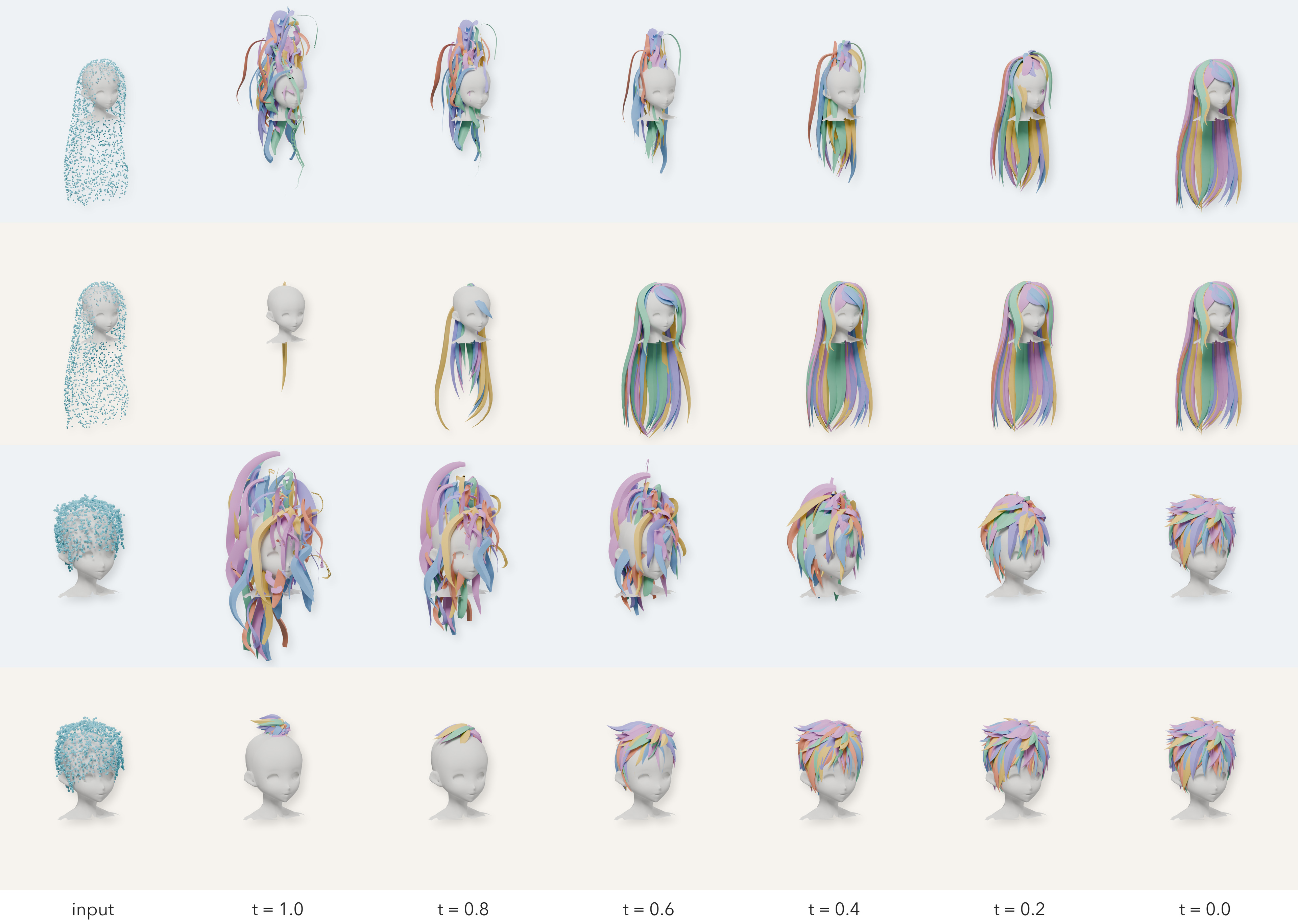}
\caption{Denoising from a point cloud. For each test point cloud (left), top: the cards decoded from the noisy token set $\mathbf{X}_t$; bottom: the clean set predicted at that step. Each card keeps its color along the trajectory.}
\label{fig:supp_traj}
\end{figure*}

\begin{figure*}[t]
\centering
\includegraphics[width=\linewidth]{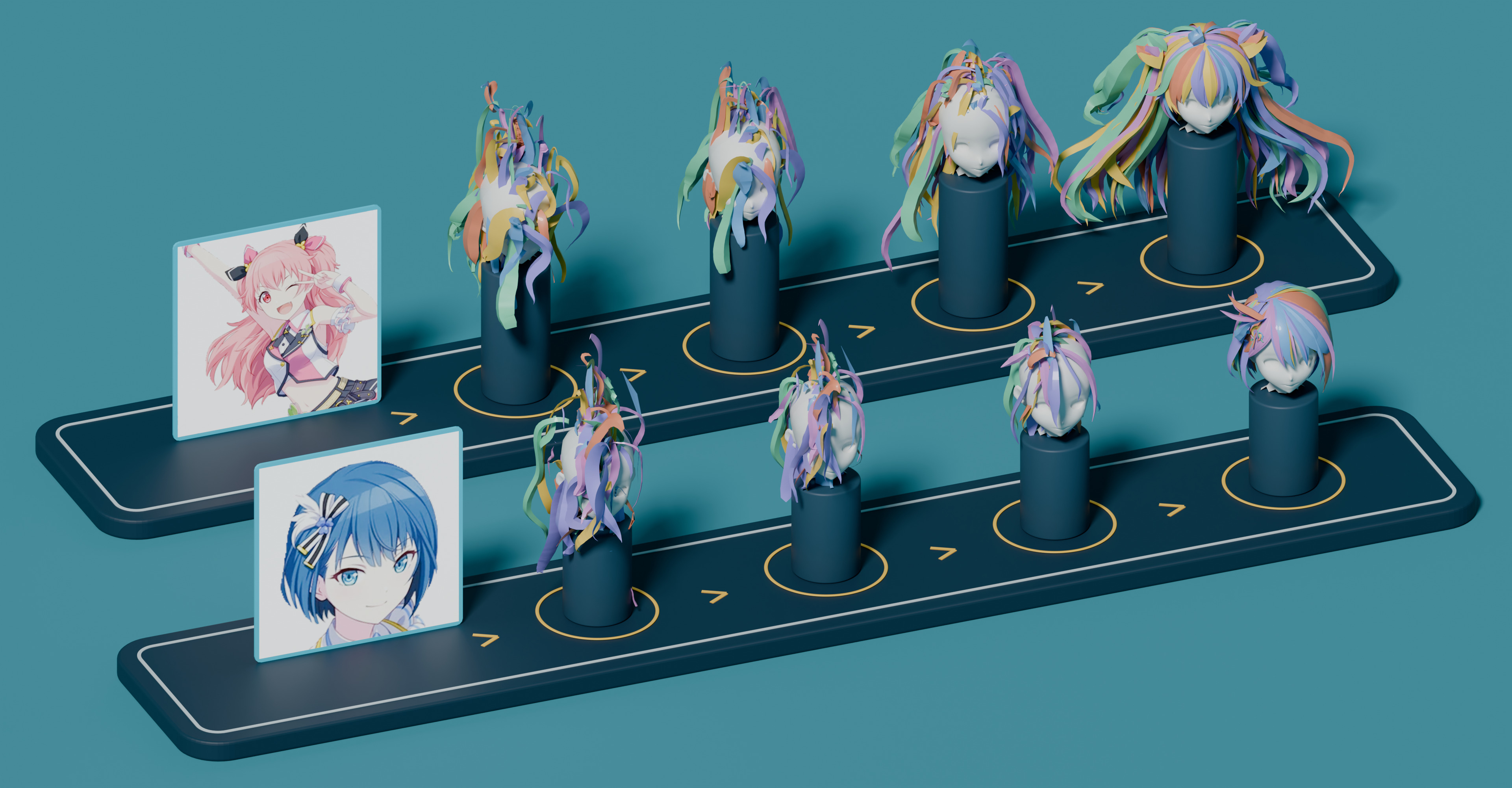}
\caption{Denoising from an illustration. On each platform, from left to right: the input, the cards decoded from the noisy token set $\mathbf{X}_t$ at $t = 1.0$, 0.7 and 0.4, and the generated hairstyle. Each card keeps its color along the trajectory.}
\label{fig:supp_traj_img}
\end{figure*}

\paragraph{Denoising.}
\Cref{fig:supp_traj} follows the set DiT through denoising from a point cloud, and \cref{fig:supp_traj_img} from an illustration.
The clean set predicted at the first steps holds only a few cards, and cards appear as their existence flags settle.

\begin{table}[t]
\centering
\caption{Drawing styles on the 76 test characters: the front view (original) and its restyles of \cref{fig:supp_style}. Aligned CD as in \cref{tab:img} and root precision as in \cref{sec:setup}. w/o restyle: without restyled images; w/o both: also without synthetic style augmentation (\cref{tab:supp_abl_img}). Bold: best model per style.}
\label{tab:supp_style}
\footnotesize
\setlength{\tabcolsep}{2.2pt}
\begin{tabular}{@{}lcccccc@{}}
\toprule
& \multicolumn{3}{c}{Aligned CD$\downarrow$} & \multicolumn{3}{c}{Root precision$\uparrow$} \\
\cmidrule(lr){2-4}\cmidrule(lr){5-7}
Style & \method & \begin{tabular}[b]{@{}c@{}}w/o\\restyle\end{tabular} & \begin{tabular}[b]{@{}c@{}}w/o\\both\end{tabular} & \method & \begin{tabular}[b]{@{}c@{}}w/o\\restyle\end{tabular} & \begin{tabular}[b]{@{}c@{}}w/o\\both\end{tabular} \\
\midrule
original & \textbf{0.0089} & 0.0090 & 0.0091 & \textbf{0.904} & 0.895 & 0.898 \\
gacha & \textbf{0.0092} & \textbf{0.0092} & 0.0093 & \textbf{0.859} & 0.830 & 0.821 \\
thick paint & \textbf{0.0098} & \textbf{0.0098} & 0.0103 & \textbf{0.818} & 0.700 & 0.676 \\
semi-real & \textbf{0.0096} & 0.0100 & 0.0102 & \textbf{0.869} & 0.812 & 0.784 \\
\bottomrule
\end{tabular}
\end{table}

\begin{figure*}[t]
\centering
\includegraphics[width=\linewidth]{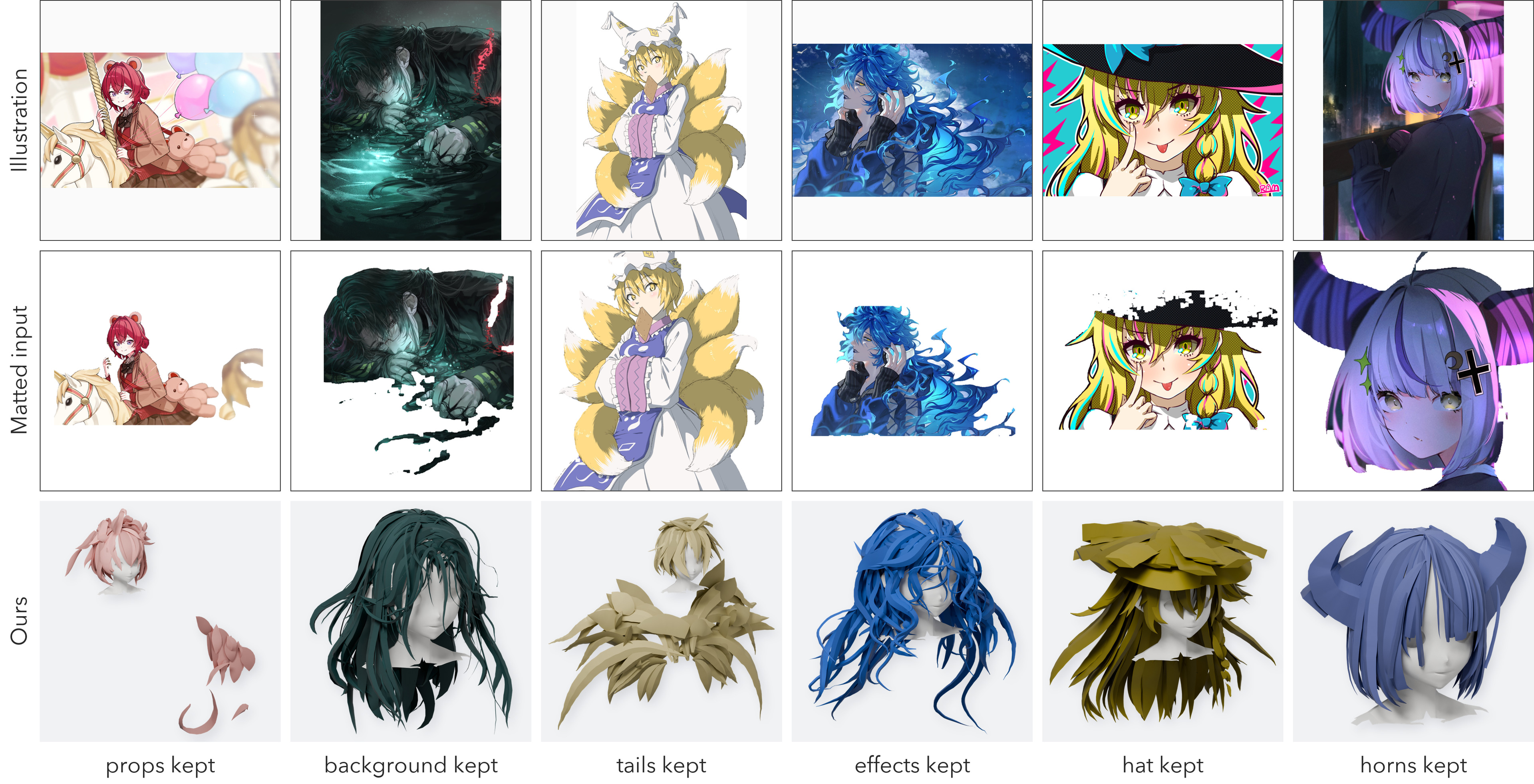}
\caption{Failure cases on in-the-wild illustrations: the input (top), the matted input (middle) and our output (bottom). The matte keeps props, background, tails, effects, a hat or horns, and they turn into cards. Hair colors are taken from the images for display.}
\label{fig:supp_fail}
\end{figure*}

\section{Failure cases}
\label{sec:supp_fail}
\Cref{fig:supp_fail} shows failure cases on in-the-wild illustrations, where objects inside the matte (\cref{sec:supp_impl}) turn into cards.

\clearpage

\end{document}